\documentclass{article}
\usepackage{graphicx}
\usepackage{amsmath}
\usepackage{amssymb}
\usepackage{booktabs}
\usepackage{multirow}
\usepackage{xcolor}
\usepackage{hyperref}
\usepackage{microtype}
\microtypesetup{expansion=false}
\usepackage{multirow}
\usepackage{authblk}
\usepackage[numbers,sort&compress]{natbib}
\usepackage[margin=1in]{geometry}

\title{Physics-Aware Self-Supervised Pretraining on Multi-Domain Raman Spectra}

\author[3]{Charalampos Davalas\thanks{Corresponding author. Email: \texttt{cdavalas@hua.gr}. ORCID: 0000-0003-4445-4314.}}
\author[3]{Georgios Levantinos\thanks{Email: \texttt{levantinos@hua.gr}.}}
\author[1,2]{Anastasios Dimitriou\thanks{Email: \texttt{a.dimitriou@inn.demokritos.gr}. ORCID: 0000-0002-1166-0800.}}
\author[1]{Nikolaos Papanikolaou\thanks{Email: \texttt{n.papanikolaou@inn.demokritos.gr}. ORCID: 0000-0002-4288-6132.}}
\author[3]{Georgios Th. Papadopoulos\thanks{Email: \texttt{g.th.papadopoulos@hua.gr}. ORCID: 0000-0003-1686-421X.}}
\author[3]{Dimitrios Michail\thanks{Email: \texttt{michail@hua.gr}. ORCID: 0000-0002-5316-6704.}}
\affil[1]{Institute of Nanoscience and Nanotechnology, NCSR ``Demokritos'', Athens, Greece}
\affil[2]{Immunoassays/Immunosensors Laboratory, Institute of Nuclear and Radiological Sciences and Technology, Energy and Safety, NCSR ``Demokritos'', Athens, Greece}
\affil[3]{Department of Informatics and Telematics, Harokopio University of Athens, Athens, Greece}
\date{}
\begin{document}
\maketitle

\begin{abstract}
Raman spectroscopy generates large volumes of unlabeled data that are difficult to exploit with conventional supervised learning. This work introduces a multi-domain self-supervised learning framework based on a masked autoencoder (MAE) for learning transferable Raman spectral representations from heterogeneous biological, chemical, forensic, and materials datasets. The proposed pretraining strategy incorporates physics-aware objectives by combining masked intensity reconstruction with a sparse Voigt-profile decoder that models spectral peak structure through differentiable peak centers, widths, and amplitudes. A denoising objective is further included to improve robustness to spectral corruption. The learned encoder is evaluated on substance classification, component identification, concentration estimation, and spectral denoising under linear probing, attention-based probing, end-to-end fine-tuning, and training from scratch. Experimental results show that self-supervised pretraining is most beneficial for challenging tasks with limited labeled data. Attention-based probing also reveals substantially more useful information in the frozen representation than linear probing, indicating that conventional linear evaluation may underestimate representation quality. In addition, the benefits of pretraining become more pronounced with increasing encoder capacity. Overall, the proposed framework demonstrates that physics-aware, multi-domain self-supervised pretraining can improve label efficiency, robustness, and transferability across diverse Raman spectroscopy tasks.
\end{abstract}

\section{Introduction}
\label{sec:intro}

Raman spectroscopy provides label-free, non-destructive molecular
fingerprints and is used routinely in chemistry, biology, materials
science, and forensics. Acquiring spectra has become cheap, as portable
spectrometers, surface-enhanced Raman scattering (SERS) substrates, and
hyperspectral imaging produce large volumes of measurements, and a single
imaging study of breast cancer cells yielded more than 1.5 million spectra
\cite{horgan2021high}. Labeling them has not become cheaper. Assigning a
spectrum to a bacterial species, a tissue type, a mineral, or a set of
mixture components requires reference measurements or expert analysis, so
labeled datasets remain small; the forensic GBL concentration dataset
used in this study, for example, contains about 1,500 spectra. Most
measured Raman spectra therefore carry no labels, and supervised models,
which learn only from the labeled part of one dataset for one task, leave
them unused.

Self-supervised learning (SSL) offers a way to use these spectra. A model
is first pretrained on unlabeled data through a pretext task and then
adapted to downstream tasks with few labels. Masked modeling, in which the
model reconstructs hidden parts of its input, has become a widely used
pretext task for language \cite{devlin2019bert} and images
\cite{he2022mae}, and has been extended to one-dimensional physiological
and time-series signals \cite{mehari2022self,zhang2024self}. For Raman
spectroscopy, SSL is at an early stage. SMAE \cite{ren2025smae} pretrains
a masked autoencoder with per-pixel reconstruction on a single bacterial
dataset and evaluates it on that same dataset, and contrastive methods
have likewise targeted bacterial classification \cite{yao2024scdc}. DSCF
\cite{xue2025deepspectral} pretrains on more than a million simulated and
experimental spectra across UV, IR, and Raman modalities, showing that
large-scale spectral pretraining is feasible. How such pretraining is best
designed for Raman spectra, and what it provides downstream, has not yet
been studied in depth.

Two properties of Raman data suggest ways to go beyond this. First,
Raman datasets from different domains share the physics that produces
their peaks while differing in how the measurement distorts them, since each
setup shifts and broadens peaks, rescales intensities, and adds its own
baseline and noise. Pretraining on a single dataset ties the encoder to
one instrument, one sample type, and one downstream task, whereas pooling
unlabeled spectra across domains exposes it to many distortions of the
same underlying physics, which may help it learn features that are less
tied to any one setup. Second, Raman bands have a recognizable shape.
Experimentally observed bands are often neither purely Gaussian nor
purely Lorentzian but contain contributions from both, and the Voigt
profile, their convolution, is commonly used to model them
\cite{sundius1973computer,meier2005art,ida2000extended}. Its center is
associated with a vibrational mode, its width with factors such as local
environment and instrumental broadening, and its amplitude with
concentration. Per-pixel reconstruction, borrowed from images, makes no
use of this structure. A pretext task that asks the encoder to describe
masked peak regions with a few Voigt profiles can therefore complement it
with a peak-level view of the spectrum. Such a description is a
parametric approximation, and a single fitted profile may stand in for
several overlapping vibrational modes, an asymmetric band, or hot-band
and intermolecular contributions, so we use it as a training signal
rather than as a claim about the underlying components
\cite{meier2005art,korepanov2018asymmetric}.

What pretraining provides downstream also remains to be characterized.
Raman analysis is not only classification, as practitioners ask which substance a
sample is, whether a given component is present in a mixture, how much of
it there is, and what the clean signal looks like beneath the noise.
Whether a single encoder serves all of these, whether its frozen features
suffice or must be adapted, and when it improves on training the same
architecture from scratch are empirical matters, and they determine
whether pretraining is worth its cost in practice.

We present a multi-domain masked autoencoder (MAE) for Raman spectra that
addresses the first two points by design and studies the third
through systematic evaluation. A
single one-dimensional Vision Transformer encoder is pretrained without
labels on ten datasets spanning biological tissue, cells and bacteria,
minerals, and chemical and forensic mixtures, several hundred thousand
spectra in total. A shared normalization of intensities and wavenumber
axes puts these datasets on a common footing, and augmentations simulate
further setup-specific distortions such as calibration shift and
stretch, intensity scaling, and detector noise. The pretext task combines three
objectives. Per-pixel reconstruction recovers the masked intensities.
Sparse Voigt reconstruction has a decoder describe each masked
peak region with a small set of Voigt profiles, predicting their centers,
widths, and amplitudes differentiably under a sparsity penalty on the
number of active profiles. Denoising trains the encoder to recover clean
spectra from inputs corrupted by Poisson shot noise and additive Gaussian
noise.

We then evaluate the pretrained encoder with the protocols standard in
the SSL literature \cite{he2022mae,chen2020simclr} on four downstream
tasks, substance classification, component identification, concentration
estimation, and spectral denoising, across twelve dataset--task pairs and
three encoder scales. Linear probing and attention probing measure what
the frozen representation already contains; end-to-end fine-tuning
measures its value as an initialization; and training the identical
architecture from random initialization isolates the contribution of
pretraining. Because the downstream datasets contribute unlabeled
training spectra to the pretraining corpus, this setting measures the
reuse of one encoder across datasets and tasks; transfer to acquisition
domains unseen during pretraining is left to future work.

We make three observations. First, pretraining helps as an initialization
where labeled data are scarce relative to the difficulty of the task.
With the base encoder, fine-tuning roughly doubles the macro-F1 of
training from scratch on 30-class bacterial identification from low
signal-to-noise spectra (0.65 vs.\ 0.33), a gain that holds at all three
scales, and it also improves forensic GBL concentration estimation (0.84
vs.\ 0.74) and R6G detection in beverages (1.00 vs.\ 0.96). Where the
downstream training set is large or the task is close to saturation, as
for the 192-class mineral subset, liver tissue, sugar mixtures, and
denoising, training from scratch catches up and pretraining is neutral or
slightly worse. Second, as reported for MAE on natural images
\cite{he2022mae}, linear probing underestimates what the pretrained
encoder contains. With the base encoder, a linear head on the frozen
features trails fine-tuning on nine of the eleven prediction tasks, and
attention pooling over the frozen token embeddings improves on linear
probing on ten of them. Third, the value of pretraining grows with
encoder scale (Appendix~\ref{app:ablations}). From the smallest to the
largest encoder, linear-probe F1 rises by 0.11 to 0.57 on seven of the
eleven prediction tasks, and the margin of fine-tuning over training from
scratch widens by 0.05 to 0.22 F1 on five of them, because fine-tuned
encoders benefit more from added capacity than encoders trained from
scratch, which degrade in four of these five cases. The margin widens
most on the 1,332-class mineral subset, where pretraining turns a
deficit with the smallest encoder into a lead with the largest, which
also surpasses the published baseline (F1 0.88, vs.\ 0.85 for
\citet{sang2022one}).

This article makes the following main contributions.
\begin{itemize}
    \item A multi-domain MAE framework that pretrains a single Raman
          spectral encoder, without labels, on ten heterogeneous datasets
          spanning biological, chemical, forensic, and materials domains,
          providing a common initialization for downstream analysis.
    \item A physics-aware pretext task that combines masked intensity
          reconstruction, sparse Voigt-profile reconstruction, and
          denoising, coupling dense signal reconstruction with an
          explicit, physically parameterized model of peak structure.
    \item A systematic evaluation of the pretrained encoder, following
          standard SSL protocols, on twelve dataset--task pairs covering
          substance classification, component identification,
          concentration estimation, and spectral denoising at three
          encoder scales, which characterizes when self-supervised
          pretraining improves on training from scratch and when it does
          not.
\end{itemize}
The source code is available at
\url{https://github.com/chdavalas/multidomain-raman-ssl}.
\section{Related Work}
\label{sec:related}

Self-supervised learning (SSL) has emerged as an effective strategy for learning transferable representations from large quantities of unlabeled data. The masked autoencoder paradigm was originally popularized by BERT \cite{devlin2019bert} for natural language processing, where randomly masked tokens are reconstructed from surrounding context. This concept was later extended to computer vision by \citet{he2022mae}, who introduced Masked Autoencoders (MAE) based on Vision Transformers (ViTs). By reconstructing heavily masked image patches, MAEs learn representations that capture global semantic structure rather than local texture statistics. Subsequent work has demonstrated that masked prediction objectives generalize effectively to one-dimensional scientific signals, including electroencephalography (EEG) \cite{banville2021uncovering,yuan2019wave2vec}, electrocardiography (ECG) \cite{mehari2022self}, and general time-series data \cite{zhang2024self}. Within Raman spectroscopy, \citet{ren2025smae} recently proposed a masked autoencoder trained on bacterial spectra using a standard per-pixel reconstruction objective, demonstrating competitive downstream classification performance under limited supervision. More broadly, foundation models for spectral data have begun to emerge beyond Raman-specific settings: \citet{xue2025deepspectral} propose a deep spectral component filtering objective as a foundation model for metabolic profiling, and \citet{yang2025spectrumworld} introduce SpectrumWorld, a general-purpose pretraining framework spanning multiple spectroscopic modalities. Neither of these approaches, nor \citet{ren2025smae}, incorporates a physically motivated decomposition of the reconstruction target. Our work extends this line of research by introducing multi-domain pretraining, physics-aware reconstruction through Voigt profile decomposition, and a comprehensive multi-task evaluation framework. Alternative SSL paradigms based on contrastive learning have also been explored \cite{yao2024scdc,chen2020simclr,grill2020byol}; however, such approaches rely heavily on carefully designed augmentations and the construction of meaningful positive and negative pairs. Reconstruction-based objectives are generally more straightforward to apply across heterogeneous spectroscopic datasets and naturally preserve spectral structure.

Deep learning has become increasingly prevalent in Raman spectroscopy, particularly for classification and identification tasks. Convolutional neural networks have been successfully applied to bacterial identification \cite{ho2019rapid}, cancer detection \cite{huang2023rapid}, pharmaceutical quality control, and materials characterization \cite{sang2022one}. More recently, transformer-based architectures have been adopted for Raman mixture quantification \cite{koyun2024ramanformer} and, in combination with convolutional feature extractors, for qualitative spectral analysis \cite{wang2024deep}. To address challenges arising from limited labeled data and instrument-specific variability, transfer learning and domain adaptation strategies have been investigated, often leveraging pretrained one-dimensional neural networks as initialization \cite{kamran2025transfer}. The availability of large-scale synthetic datasets has further expanded opportunities for representation learning. For example, \citet{liang2025dataset} introduced a dataset containing more than 220,000 density functional theory (DFT)-computed Raman and infrared spectra derived from ChEMBL molecules, providing a complementary source of training data beyond experimentally acquired measurements. In the self-supervised setting, \citet{yao2024scdc} proposed a dual-perspective contrastive framework with self-calibration mechanisms for bacterial Raman classification under semi-supervised conditions, highlighting the growing interest in SSL approaches for spectroscopic analysis.

A parallel line of research focuses on incorporating physical knowledge into machine learning models. Physics-informed neural networks (PINNs) integrate governing physical principles as constraints during optimization, improving generalization and reducing dependence on large labeled datasets \cite{raissi2019pinn}. Within spectroscopy, \citet{vemuri2025rampinn} embedded the Kramers--Kronig relations as differentiable constraints to recover Raman spectra from coherent anti-Stokes Raman scattering (CARS) measurements, demonstrating that physically motivated inductive biases can facilitate zero-shot transfer from synthetic to experimental data. In a related direction, \citet{georgiev2024hyperspectral} used physics-constrained autoencoders to perform hyperspectral unmixing of Raman spectra, encoding chemical mixture priors directly into the network architecture; unlike our approach, this physical constraint is imposed at the level of a supervised downstream task rather than a pretraining objective. Another longstanding challenge in spectroscopy is the decomposition of spectra into physically meaningful peak profiles. Voigt and pseudo-Voigt profiles are widely used to fit Raman bands, typically by nonlinear least-squares optimization \cite{sundius1973computer,meier2005art,ida2000extended}, although symmetric profiles can be unreliable for asymmetric or overlapping bands \cite{korepanov2018asymmetric}. Deep learning has also been explored to accelerate peak fitting in other spectroscopies, for example in two-dimensional NMR \cite{li2021deep}. However, existing methods typically treat peak fitting as a standalone downstream task. To the best of our knowledge, incorporating sparse Voigt decomposition directly into a self-supervised reconstruction objective has not previously been investigated.

The evaluation of pretrained representations has become increasingly standardized within the SSL literature. Linear probing and end-to-end fine-tuning are widely used to assess the quality and transferability of learned representations \cite{he2022mae,chen2020simclr}. Linear probing evaluates the discriminative power of frozen features, whereas fine-tuning measures the extent to which pretrained representations can be adapted to a downstream task. Comparisons against models trained from random initialization provide a direct measure of the sample-efficiency gains attributable to pretraining, particularly under limited-label settings.

Standardized benchmarking of machine learning methods for Raman spectroscopy has itself only recently begun. Two concurrent efforts have taken initial steps toward unified evaluation protocols, albeit at more limited scale: \citet{sineesh2026benchmarking} compare five Raman-specific deep learning classifiers across three open-source datasets, and \citet{lange2025benchmarking} benchmark eleven tabular and deep learning models, including TabPFN, on a single bioprocess-monitoring regression dataset. \citet{koddenbrock2026ramanbench} substantially extend both efforts and introduced RamanBench, a large-scale, living benchmark unifying 74 public Raman datasets (163 prediction targets, 325,668 spectra) spanning material science, biotechnology, medical, and chemical/industrial domains, and evaluating 28 model families --- from classical chemometrics (PLS, SVM) and gradient-boosted trees to Raman-specific deep architectures, time-series classifiers (ROCKET, Arsenal), and Tabular Foundation Models (TabPFN, TabICL, MITRA) --- under a standardized protocol with Elo-rating, normalized-score, and improvability metrics. Their central findings are that Tabular Foundation Models and time-series classifiers consistently outperform Raman-specific architectures and gradient-boosting baselines, that Partial Least Squares remains a surprisingly durable long-standing baseline, and that no single model family generalizes across all datasets. Notably, RamanBench evaluates every model as a supervised learner trained from scratch on each individual prediction target; it does not consider self-supervised pretraining, cross-domain representation transfer, or physics-aware reconstruction objectives. Our work is complementary to this effort: rather than proposing a new supervised architecture to compete on RamanBench's leaderboard, we ask whether a single encoder, pretrained once via physics-aware masked reconstruction across heterogeneous unlabeled Raman corpora, can transfer across exactly this kind of multi-domain, multi-task diversity --- potentially narrowing the generalization gap that RamanBench identifies.

\section{Datasets}
\label{sec:datasets}

We assemble a multi-domain pretraining corpus from the following
experimental Raman datasets, spanning diverse biological, chemical,
forensic, and materials domains. For a compact overview of the corpus,
including split sizes and spectral characteristics, see
Table~\ref{tab:datasets}.

\paragraph{Adenine mixtures.}
The adenine dataset \cite{fornasaro2020surface} consists of Raman spectra acquired from pure adenine samples at multiple concentration levels and under varying acquisition conditions. This dataset comprises the results of the first interlaboratory study (ILS) focused on the quantitative application of surface-enhanced Raman scattering (SERS). It was generated within the framework of Working Group 1 (WG1) of the EU COST Action BM1401 (Raman4Clinics) and involves contributions from 15 independent laboratories and 44 researchers. The dataset was designed to systematically evaluate the reproducibility and trueness of SERS-based quantitative measurements across different experimental conditions. Specifically, the dataset includes measurements obtained using diverse SERS methodologies, reflecting variations in instrumentation, substrates, analytes, and analytical matrices. This diversity enables a comprehensive assessment of method performance and interlaboratory variability. In addition, the dataset supports comparative analysis between different quantitative SERS approaches, highlighting factors that influence accuracy and robustness. Overall, this dataset serves as a benchmark resource for the validation and standardization of SERS protocols. By providing a community-driven reference for evaluating analytical performance, it contributes to improving the reliability and broader acceptance of SERS in regulated and application-oriented environments. The dataset was converted into a binary classification task by thresholding the concentration values to distinguish samples below and above a defined cutoff.

\paragraph{Bacteria.}
The bacterial dataset \citet{ho2019rapid, ren2025smae} consists of Raman spectra acquired from 30 clinically relevant pathogenic bacterial isolates, with each species exhibiting distinct yet partially overlapping biochemical signatures associated with proteins, lipids, nucleic acids, and other cellular constituents. Originally introduced by \citet{ho2019rapid}, the dataset was developed to evaluate rapid pathogen identification using Raman spectroscopy combined with deep learning approaches. Spectra were collected under low signal-to-noise conditions, reflecting realistic experimental variability and enabling the assessment of robust representation learning methods for microbial classification. Due to substantial intra-class variability, biological heterogeneity, and fluorescence interference, the dataset provides a challenging benchmark for evaluating the robustness and generalization ability of machine learning models for Raman-based bacterial identification.

\paragraph{Liver Cancer.}
This dataset \cite{huang2023rapid} comprises Raman spectroscopic measurements of human hepatic tissue samples, developed to support pathological diagnosis of liver carcinoma. It includes spectral data acquired from both carcinoma tissues and adjacent non-tumorous tissues, enabling comparative analysis for accurate tissue classification. The dataset is generated using a label-free, non-destructive Raman spectroscopy workflow and is designed for both in vitro and intraoperative diagnostic applications. It incorporates high-resolution spectral and imaging data, including two-dimensional (2D) and three-dimensional (3D) Raman maps of unprocessed tissue sections at submicrometric resolution. These data capture detailed molecular composition, facilitating the identification of tumor boundaries and structural heterogeneity. In addition to binary classification (tumor vs. non-tumor), the dataset contains annotations supporting advanced pathological characterization, such as cancer subtype, differentiation grade, and tumor stage. Although the data readily available focus more on binary classification between Cancerous and Paracancerous tissues (Binary classification). Validation is further supported through complementary metabolomics data, enhancing the reliability of biochemical interpretations. Overall, this dataset provides a comprehensive resource for the development and evaluation of machine learning and deep learning models aimed at rapid, real-time liver cancer diagnosis, including potential deployment in intraoperative settings using portable Raman systems.

\begin{table}[t]
\centering
\small
\renewcommand{\arraystretch}{1.15}
\setlength{\tabcolsep}{5pt}
\caption{\textbf{Multi-domain pretraining corpus.} Domain, whether
windowed sampling is applied, resulting train/validation/test spectrum
counts, and spectral range and resolution (cm$^{-1}$) for each dataset;
paired rows show the effect of windowed sampling on split size and
resolution. All spectra are used unlabeled during pretraining.}
\vspace{4pt}
\label{tab:datasets}
\begin{tabular}{l c l r c c}
\toprule
\textbf{Dataset} & \textbf{Windowed} & \textbf{Domain} & \textbf{\# Spectra} & \textbf{Range} & \textbf{Resolution} \\
 & \textbf{Sampling} & & \textbf{(Train/Val/Test)} & (cm$^{-1}$) & (cm$^{-1}$) \\
\midrule
\multirow{2}{*}{Adenine mixtures} & \textcolor{red!80!black}{$\times$} & \multirow{2}{*}{Biological} & 2107/234/1175 & \multirow{2}{*}{400--1999} & 3 \\
 & \textcolor{green!60!black}{$\checkmark$} & & 6321/702/3525 & & 6 \\
\cmidrule(lr){1-6}
\multirow{2}{*}{Bacteria} & \textcolor{red!80!black}{$\times$} & \multirow{2}{*}{Biological} & 48000/12000/3000 & \multirow{2}{*}{381--1792} & $\approx 1.4$ \\
 & \textcolor{green!60!black}{$\checkmark$} & & 192000/24000/24000 & & $\approx 2.8$ \\
\cmidrule(lr){1-6}
\multirow{2}{*}{Liver Cancer} & \textcolor{red!80!black}{$\times$} & \multirow{2}{*}{Biological} & 4000/1000/1000 & \multirow{2}{*}{500--2000} & $\approx 1.68$ \\
 & \textcolor{green!60!black}{$\checkmark$} & & 16000/4000/4000 & & $\approx 3.36$ \\
\cmidrule(lr){1-6}
\multirow{2}{*}{RRUFF-S 1\&2} & \textcolor{red!80!black}{$\times$} & \multirow{2}{*}{Minerals} & 9570/139/4161 & \multirow{2}{*}{30--1599} & 1 \\
 & \textcolor{green!60!black}{$\checkmark$} & & 57422/832/24966 & & 2 \\
\cmidrule(lr){1-6}
R6G & \textcolor{red!80!black}{$\times$} & Component identification & 56084/7010/7010 & 531--1821 & $\approx 1.08$ \\
\cmidrule(lr){1-6}
Raman sugars & \textcolor{red!80!black}{$\times$} & Component identification & 6272/784/784 & 142--3684 & $\approx 1.77$ \\
\cmidrule(lr){1-6}
GBL & \textcolor{red!80!black}{$\times$} & Concentration estimation & 1207/150/150 & 513--1636 & $\approx 1.08$ \\
\cmidrule(lr){1-6}
DeepRaman & \textcolor{red!80!black}{$\times$} & Component identification & 3510/390/36 & 50--3200 & 2 \\
\cmidrule(lr){1-6}
Hyperspectral & \textcolor{red!80!black}{$\times$} & Reconstruction & 141330/17666/17666 & 498--1799 & $\approx 2.6$ \\
\cmidrule(lr){1-6}
MDA-Spectral & \textcolor{red!80!black}{$\times$} & Denoising & 159618/15961/12694 & 498--1799 & $\approx 2.6$ \\
\midrule
\multicolumn{3}{l}{\textbf{Total}} & \textbf{639764/71495/94831} & & \\
\bottomrule
\end{tabular}
\end{table}

\paragraph{RRUFF Subset (RRUFF-S)}
This subset of mineral data, from the work of~\cite{sang2022one}, is derived from the RRUFF \cite{lafuente20151} Raman spectral database and consists of a large collection of mineral Raman spectra curated for multi-class classification tasks. It includes spectral measurements from a wide variety of mineral samples, encompassing hundreds to potentially thousands of distinct classes with varying sample sizes per class. To support comprehensive evaluation, the dataset is organized into two sub-datasets, each designed to assess classification performance under different conditions. These sub-datasets enable analysis of model robustness across scenarios involving large class diversity and imbalanced class distributions, without the application of data augmentation techniques. The dataset captures detailed molecular information inherent to mineral compositions, making it suitable for benchmarking both traditional machine learning approaches (e.g., support vector machines and k-nearest neighbors) and deep learning architectures. In particular, it is well-suited for evaluating one-dimensional convolutional neural networks (1D CNNs) for spectral classification tasks. Overall, this dataset provides a challenging and scalable benchmark for high-dimensional, multi-class Raman spectral classification, supporting the development and validation of advanced deep learning models capable of handling large-scale and imbalanced datasets with high accuracy. In the paper, the dataset has been evaluated in two different parts one with 192 classes and the second part which has an larger number of classes 1332 and is a more demanding set in terms of substance classification. 

\paragraph{R6G (Rhodamine 6G).}
R6G is a highly fluorescent dye commonly employed as a tracer and laser dye. The R6G dataset, acquired through measurements conducted at the National Centre for Scientific Research ``Demokritos,'' is used for the component identification task, where the objective is to determine the presence or absence of a target substance. It comprises positive mixtures, in which R6G was present in solvents such as water, Coca-Cola, and whiskey; negative mixtures obtained from beverages including beer, rum, Fanta, red and white wine, and vodka; and reference spectra of pure R6G measured on different substrates.

\paragraph{Raman sugars.}
This dataset by \cite{raman_sugars}, comprises a comprehensive collection of Raman spectra acquired from controlled mixtures of aqueous sugar solutions, and is intended as a benchmark resource for the development and evaluation of spectral analysis methodologies. The dataset includes measurements from 245 unique mixtures composed of four sugars—sucrose, fructose, maltose, and glucose—combined with water in a full factorial design, enabling systematic variation in concentration profiles. Two complementary sub-datasets are provided, which correspond to identical sample compositions measured under different acquisition conditions. Specifically, these datasets differ in integration time (5 s and 0.5 s, respectively), resulting in distinct signal-to-noise ratio (SNR) characteristics. Each mixture is measured multiple times to capture experimental variability, yielding 1,960 spectra in the high-SNR dataset and 7,840 spectra in the low-SNR dataset. The inclusion of repeated measurements, structured into acquisition rounds and repetitions, allows for the assessment of temporal effects and measurement consistency. Overall, this dataset provides a high-quality, well-annotated, and experimentally rigorous benchmark for quantitative Raman spectroscopy, supporting tasks such as concentration estimation, noise robustness analysis, and the development of advanced machine learning and deep learning models for spectral interpretation.

\paragraph{GBL (Gamma-Butyrolactone).}
GBL is a forensically relevant solvent and controlled-substance precursor whose quantification in seized samples represents a practical analytical challenge. The GBL dataset, acquired through measurements conducted at the National Centre for Scientific Research ``Demokritos,'' is used for the concentration estimation task, formulated as a multi-class classification problem over discrete concentration bins rather than as continuous regression. It comprises Raman spectroscopic measurements of samples containing varying concentrations of gamma-butyrolactone (GBL), an industrial solvent that is rapidly converted into gamma-hydroxybutyrate (GHB) following ingestion. The dataset was designed to support classification and detection across concentration levels relevant to analytical and forensic applications. Specifically, it includes four discrete GBL concentrations: 0 mg/mL, 0.1 mg/mL, 1 mg/mL, and 10 mg/mL. Overall, the dataset enables the development and validation of spectral analysis methods for both binary presence-or-absence detection and multi-class concentration classification, particularly under conditions of low analyte concentration and uneven class distribution.

\paragraph{Hyperspectral dataset.}
The Hyperspectral dataset \cite{horgan2021high} consists of high-resolution, spatially resolved hyperspectral Raman images acquired from in vitro MDA-MB-231 human breast cancer cells. Each sample provides a complete spatial-spectral data cube that captures the complex molecular signatures of intricate cellular structures, such as the nucleus and lipid droplets, offering a highly detailed and uncorrupted representation of the biological samples. This dataset comprises a large-scale collection of hyperspectral Raman imaging data designed to support the development of deep-learning methods for accelerated molecular imaging. It contains over 1.5 million Raman spectra, corresponding to approximately 400 hours of experimental acquisition, and represents one of the most extensive datasets for high-throughput Raman spectroscopy applications.

\paragraph{MDA-Spectral dataset.}
The MDA-Spectral dataset is also included in the Hyperspectral dataset \cite{horgan2021high} and comprises spectra acquired from in vitro MDA-MB-231 human breast cancer cells. The dataset consists of paired short-acquisition, low signal-to-noise ratio (SNR) spectra and corresponding long-acquisition, high SNR target spectra. This dataset is used for the spectral denoising task, aiming to digitally recover clean molecular signatures from highly noisy inputs without sacrificing chemical information. The dataset is structured to enable multiple computational tasks, including spectral denoising, image reconstruction, and spatial super-resolution of hyperspectral Raman images. It captures molecular signatures with varying signal-to-noise ratios, enabling the training and evaluation of models aimed at improving spectral quality and imaging efficiency. In particular, it supports the recovery of clean Raman signals from noisy measurements and the enhancement of spatial resolution by factors of 2–4× while preserving biologically relevant molecular information.

\paragraph{DeepRaman dataset.}
Originally curated for training the DeepRaman component identification model \cite{fan2023universal}, this composite dataset relies on distinct subsets for its original training and evaluation phases. The foundational training corpus comprises a comprehensive collection of standardized Raman spectra, aggregating broad multi-wavelength measurements (1064 nm, 785 nm, 532 nm, and 488 nm) from the extensive S.T. Japan database, alongside a targeted subset of pure pharmaceutical compounds acquired via a B\&W Tek i-Raman-785S spectrometer. As detailed in the original study, this training data was expanded through a synthetic augmentation strategy to generate positive and negative spectral pairs by combining pure component spectra with simulated complex mixtures. To increase spectral diversity, these simulated mixtures were generated using randomized concentration ratios and subjected to stochastic peak shifting, noise injection, and intensity scaling, followed by a final maximum-value normalization. Conversely, for testing, the original study utilized two experimental sets: a multi-component dataset comprising of 105 physical mixtures of up to 13 common solvents, and a liquid and powder dataset consisting of 6 experimental mixture samples. In our current study, we maintain this functional separation: the extensive augmented training corpus is utilized exclusively for the pretraining of the MAE, while the experimental multi-component and liquid and powder datasets are reserved for evaluating the downstream component identification task.

\section{Methodology}
\label{sec:method}

\begin{figure}[t]
    \centering
    \includegraphics[width=1\linewidth]{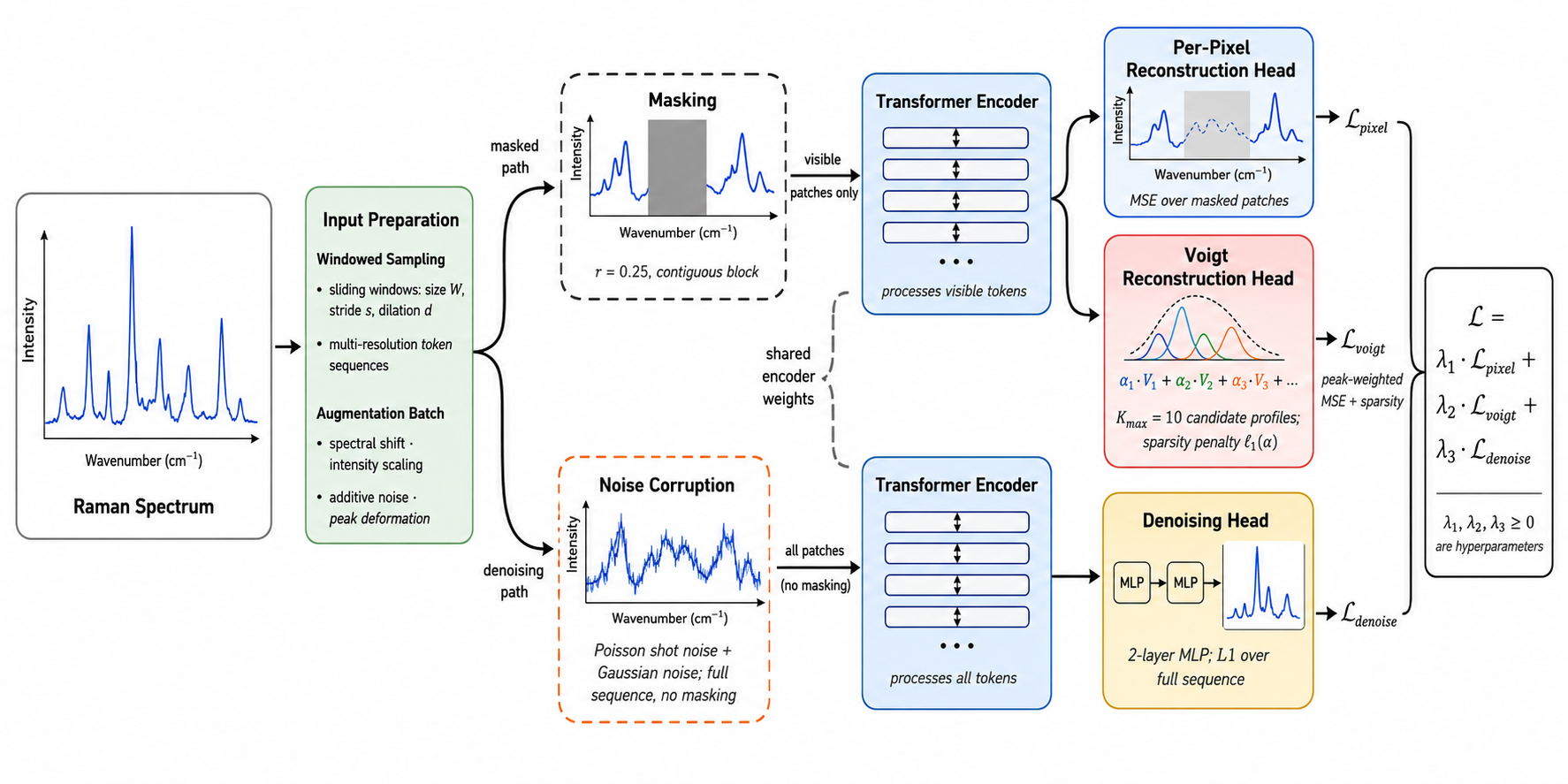}
    \caption{\textbf{Pretraining architecture.} The input spectrum is
    tokenized into patches and partially masked; a Transformer encoder
    processes the visible tokens and a decoder reconstructs the masked
    regions through three heads (per-pixel reconstruction, sparse
    Voigt-profile reconstruction, and denoising).}
    \label{fig:architecture}
\end{figure}

This section describes the pipeline used to pretrain and instantiate our
Raman spectra foundation model. Our pretraining framework follows the
masked autoencoder (MAE) paradigm, in which a fraction of the input spectrum is
masked, a Transformer encoder processes the visible tokens, and a decoder
reconstructs the masked regions. We depart from standard MAE in three
ways, namely (1) a physics-aware Voigt reconstruction head that produces
interpretable peak decompositions, (2) an auxiliary denoising objective,
and (3) a multi-objective loss combining these components. The encoder is
a Vision Transformer (ViT) adapted to one-dimensional spectral inputs and
processes only the visible tokens during masked reconstruction; a decoder
then reconstructs the masked regions. Figure~\ref{fig:architecture}
provides an overview of the full architecture. Three model scales are
evaluated --- MAE-S, MAE-B, and MAE-L --- with corresponding encoder and
decoder depths summarized in Table~\ref{tab:model-scales}. Positional
information is incorporated through either trainable embeddings, fixed
sinusoidal embeddings, or physically-informed Fourier embeddings derived
from the Raman wavenumber axis. During pretraining no [CLS] token is
used, whereas for downstream tasks a [CLS] token is prepended and its
final embedding is used as the spectrum representation.

\begin{table}
    \centering
    \caption{\textbf{Model-scale configurations.} Encoder/decoder
    dimensions for the three evaluated model sizes (MAE-S, MAE-B, MAE-L):
    embedding size, encoder and decoder depth, MLP hidden size, and
    attention head count and size.}
    \vspace{4pt}
    \begin{tabular}{ccccccc}
    \hline
    Model size & embedding size &  encoder depth & decoder depth & mlp size & \# heads & head size \\
    \hline
    \hline
    \textbf{MAE-S} (Small) & 64 & 4 & 2 & 256 & 4 & 64\\
    \hline
    \textbf{MAE-B} (Base)  & 128 & 4 & 2 & 512 & 4 & 128 \\
    \hline
    \textbf{MAE-L} (Large) & 256 & 6 & 2 & 1024 & 4 & 256 \\
    \hline
    \end{tabular}
    \label{tab:model-scales}
\end{table}

We first describe the preprocessing pipeline shared by all datasets
(Section~\ref{sec:preprocessing}), then the masking strategy used during
pretraining (Section~\ref{sec:masking}), and finally the multi-objective
reconstruction and denoising losses that define our pretraining objective
(Section~\ref{sec:objectives}).

\subsection{Preprocessing}
\label{sec:preprocessing}

Prior to tokenization and model input, all Raman spectra undergo a standardized preprocessing pipeline. Let $\mathbf{x} \in \mathbb{R}^{T}$ denote the raw intensity spectrum of a given sample, together with its associated wavenumber axis; when explicit wavenumber values are not available, they are reconstructed from the Raman shift information reported in the original data source. Both the wavenumber axis and the intensity values are normalized via global min-max scaling computed over the entire dataset corpus, mapping values to $[0,1]$. This ensures comparability across datasets acquired under different experimental conditions and mitigates the influence of outliers and numerical artifacts; datasets already provided within a fixed numerical range are left unchanged.

\paragraph{Tokenization.} Following normalization, spectra are converted into patch-level representations using a sliding-window procedure parameterized by a window size $W$, stride $s$, and dilation factor $d$, the latter controlling the effective sampling resolution within each window. By default $W$ spans the entire spectrum, yielding a single global token. Window extraction is implemented using strided views to avoid unnecessary memory copies, after which the windows are materialized as a contiguous tensor for subsequent embedding and Transformer processing.

\paragraph{Data augmentation.} To improve robustness and reduce sensitivity to acquisition-specific artifacts, several spectral augmentations are applied independently during self-supervised pretraining. Spectral shift augmentation introduces a discrete random displacement of the spectral axis by an integer offset $k \sim \mathcal{U}\{-k_{\max},\dots,k_{\max}\}$ using circular rolling, simulating small calibration misalignments between measurements. Intensity scaling multiplies the entire spectrum by a random factor $s \sim \mathcal{U}(s_{\min}, s_{\max})$, reflecting variations in laser power, detector sensitivity, and sample concentration. Additive Gaussian noise,

\[
\mathbf{x} + \boldsymbol{\epsilon},
\qquad
\boldsymbol{\epsilon} \sim \mathcal{N}(\mathbf{0}, \sigma^2 \mathbf{I}),
\]

with $\sigma$ proportional to the spectrum's standard deviation, models measurement noise and sensor uncertainty. Finally, peak shift and stretch augmentation applies a continuous deformation of the spectral axis using a random shift $\delta \sim \mathcal{U}(\delta_{\min}, \delta_{\max})$ and a stretch factor $\alpha \sim \mathcal{U}(\alpha_{\min}, \alpha_{\max})$, followed by linear interpolation, simulating non-uniform calibration drift and small physical variations in peak positions commonly encountered in Raman spectroscopy.

\subsection{Masking Strategies}
\label{sec:masking}

Given an input spectrum divided into $T=P$ patches, we mask a fraction $r$
of the sequence during pretraining. The number of masked patches is
$M=\lfloor rT \rfloor$, and the masked and visible patch sets are denoted
by $\mathcal{M}\subset\{0,\dots,T-1\}$ and
$\mathcal{U}=\{0,\dots,T-1\}\setminus\mathcal{M}$, respectively. We
consider four masking strategies.

\paragraph{Random contiguous masking.}
A contiguous block of $M$ patches is removed at a random position. The
starting index is sampled as $s \sim \mathrm{Uniform}\{0,\dots,T-M\}$, so
that $\mathcal{M}=\{s,s+1,\dots,s+M-1\}$. Because the block is contiguous,
the model must infer the missing structure from broader context rather
than interpolate from adjacent patches.

\paragraph{Peak-centered masking.}
The dominant peak is located at $p=\arg\max_t x_t$, and the mask is a
window around it, $\mathcal{M}=\{p-M_{\mathrm{left}},\dots,p+M_{\mathrm{right}}-1\}$
with $M_{\mathrm{left}}+M_{\mathrm{right}}=M$. If the window extends
beyond the spectral boundaries, it is shifted to remain within
$\{0,\dots,T-1\}$. This forces the model to reconstruct the most
informative region of the spectrum.

\paragraph{Voigt peak masking.}
This strategy uses the same peak-centered indices, but the masked region
is assigned to the physics-aware Voigt reconstruction objective rather
than to per-pixel reconstruction alone.

\paragraph{Front masking.}
The final contiguous region of the spectrum is removed, so that
$\mathcal{M}=\{T-M,\dots,T-1\}$ and $\mathcal{U}=\{0,\dots,T-M-1\}$. The
model must then reconstruct the missing region using only the preceding
visible context.

\subsection{Reconstruction and Denoising Objectives}
\label{sec:objectives}

\paragraph{Per-pixel reconstruction.}

The first reconstruction branch is a per-pixel decoder that predicts
the original spectral values for all patches. The reconstruction loss
over masked patches is

\begin{equation}
    \mathcal{L}_{\text{pixel}} = \frac{1}{|M|} \sum_{i \in M}
    \bigl\| \hat{\mathbf{x}}_i - \mathbf{x}_i \bigr\|_2^2,
\end{equation}
where $M$ denotes the masked patch indices and
$\hat{\mathbf{x}}_i$ is the decoder output for patch $i$.

\paragraph{Physics-aware Voigt reconstruction.}

To incorporate physically meaningful peak representations, a second
reconstruction branch predicts sparse Voigt-profile decompositions.
A Voigt profile is defined as the convolution of Lorentzian and Gaussian
functions, a more detailed description of the loss functions is included
in Appendix~\ref{app:voigt},

The Voigt head predicts up to $K_{\max}$ candidate profiles for each
masked patch. Each profile is parameterized by
$\boldsymbol{\theta}_k^{(i)}=(A_k,\nu_k,\Gamma_L^k,\Gamma_G^k)$ and a
soft presence score $\alpha_k^{(i)}$. The reconstructed spectrum is
formed as a weighted sum of candidate profiles, while positivity and
physical validity are enforced through softplus and sigmoid
parameterizations.The corresponding reconstruction objective combines peak-weighted
reconstruction, global reconstruction, and sparsity regularization.

\paragraph{Denoising objective.}

In addition to masked reconstruction, we introduce a denoising pretext
task. A corrupted spectrum $\tilde{\mathbf{x}}$ is generated using
Poisson shot noise and additive Gaussian noise. Unlike the masking
objectives, the encoder processes the complete unmasked sequence.
A dedicated two-layer MLP predicts the clean spectrum, and the loss is

\begin{equation}
\mathcal{L}_{\text{denoise}} =
\frac{1}{N} \sum_i \left| f_{\theta}(\tilde{\mathbf{x}})_i - x_i \right|
\end{equation}
where $f_{\theta}$ denotes the encoder-decoder network.

\paragraph{Overall pretraining objective.}

The overall pretraining objective jointly optimizes the reconstruction
and denoising tasks:

\begin{equation}
    \mathcal{L}_{\text{pretrain}} =
    \lambda_1\,\mathcal{L}_{\text{pixel}} +
    \lambda_2\,\mathcal{L}_{\text{Voigt}} +
    \lambda_3\,\mathcal{L}_{\text{denoise}},
    \label{eq:total_loss}
\end{equation}
where $\lambda_1$, $\lambda_2$, and $\lambda_3$ are non-negative
hyperparameters. Setting $\lambda_2=0$ recovers a standard pixel-MAE,
whereas setting $\lambda_1=0$ yields a purely physics-aware model.

\section{Experiments}
\label{sec:experiments}

This section describes how we assess the quality and downstream utility
of the representations learned during pretraining. We first define the
four downstream tasks used to probe the pretrained encoder, spanning the
primary analytical objectives of Raman spectroscopy
(Section~\ref{sec:tasks}). We then describe four complementary evaluation
protocols that adapt the frozen or pretrained encoder to each task,
probing different aspects of the learned representations from
frozen-feature quality to full fine-tuning (Section~\ref{sec:eval}).
Section~\ref{sec:results} reports the resulting performance across all
task/protocol combinations.

\subsection{Downstream Tasks}
\label{sec:tasks}

We evaluate the learned representations on four downstream tasks that collectively capture the primary analytical objectives of Raman spectroscopy, including classification, component identification, concentration estimation, and signal reconstruction.

The first task is substance classification, where the objective is to assign a Raman spectrum to one of a predefined set of classes. This task is implemented using a ViT1D encoder followed by a classification head. Depending on the dataset, the problem may be formulated as either binary classification, where the model distinguishes between two classes, or multi-class classification, where the model predicts the correct label among $C$ molecular species, bacterial strains, or chemical categories. This setup follows established Raman spectroscopy benchmarks, including bacterial identification datasets in which each spectrum is directly associated with a known class \cite{ho2019rapid}.

The second task is component identification. Given a query spectrum and a reference spectrum corresponding to a candidate component, the model must determine whether the reference component is present in the query. Rather than predicting a class label from a single spectrum, this task is formulated as a similarity-learning problem. We employ a pseudo-Siamese architecture in which the query and reference spectra are processed by separate encoder branches and projected into a shared embedding space. Training is performed using positive and negative spectrum pairs, encouraging samples containing the same component to produce similar embeddings while separating representations associated with different components. This formulation supports both binary component detection and multi-component identification by comparing a query spectrum against a library of reference spectra. Similar reference-based matching strategies have previously been explored for Raman spectral analysis \cite{fan2023universal}.

The third task is concentration estimation. Given a spectrum from a sample of known composition but unknown concentration, the model predicts the concentration level as one of $C$ discrete classes. This formulation reflects practical analytical settings in which concentrations are assigned to calibrated ranges (e.g., trace, low, medium, and high) rather than estimated as continuous values. Such scenarios arise in applications including forensic quantification of GBL and quantitative SERS analysis using R6G. Concentration bins are defined separately for each dataset according to the physical concentration ranges available during data collection. Performance is evaluated using classification accuracy and macro-averaged F1 score.

The final task is spectral denoising, where the model reconstructs a clean Raman spectrum from a noisy observation. Unlike classification-based tasks, denoising directly evaluates the extent to which the learned representations capture underlying spectral structure and peak morphology beyond class-discriminative information \cite{horgan2021high}.Denoising is evaluated in two settings: (i) fine-tuning, in which the pretrained encoder initializes a denoising network that is subsequently optimized on task-specific data, and (ii) training from scratch, in which the same denoising architecture is randomly initialized and trained on
the same task-specific data. Performance is assessed using Mean Squared Error (MSE), Peak Signal-to-Noise Ratio (PSNR) and the Structural Similarity Index Measure (SSIM).

Table~\ref{tab:dataset_per_task} summarizes the downstream task assigned
to each dataset. For substance classification, the label count denotes
the number of substance or sample classes, whereas for concentration
estimation it denotes the number of discrete concentration levels.
Component identification is formulated as a binary presence-or-absence
decision for a specified target component in a mixture. Thus, the value
of one reported for R6G and Raman Sugars denotes one target component,
rather than one output class: each prediction distinguishes a negative
sample from a sample containing R6G or the target sugar, respectively.
Spectral denoising has no discrete label space, as the target is a
high-SNR spectrum.

\begin{table}
\centering 
\caption{\textbf{Dataset-to-task mapping.} Downstream task assignment for each dataset, together with the number of labels(classes, number of components, or concentration levels) defining the corresponding label space.}
\label{tab:dataset_per_task} 
\begin{tabular}{crc} 
\hline 
\textbf{Task} & \textbf{Dataset} & \textbf{\# Labels} \\
\hline 
\multirow{5}{*}{Classification} & Adenine  & 2 \\
                                & Bacteria & 30 \\
                                & Liver Cancer & 2 \\
                                & RRUFF-S (1/2) & 192 \\
                                & RRUFF-S (2/2) & 1332 \\
\hline 
\multirow{4}{*}{Component Identification} & DeepRaman(Multi-Component) &  13 \\
                                          & DeepRaman(Liquid \& Powder) & 6 \\
                                          & R6G & 1 \\
                                          & Raman Sugars & 1\\
\hline 
\multirow{2}{*}{Concentration Estimation} & GBL & 4 \\
                                          & Raman Sugars & 5 \\
\hline 
Spectral Denoising & MDA-Spectral & -- \\ 
\hline 
 
\end{tabular} 
\end{table}

\subsection{Evaluation Protocols}
\label{sec:eval}

\begin{figure}[h]
    \centering
    \includegraphics[width=0.9\linewidth]{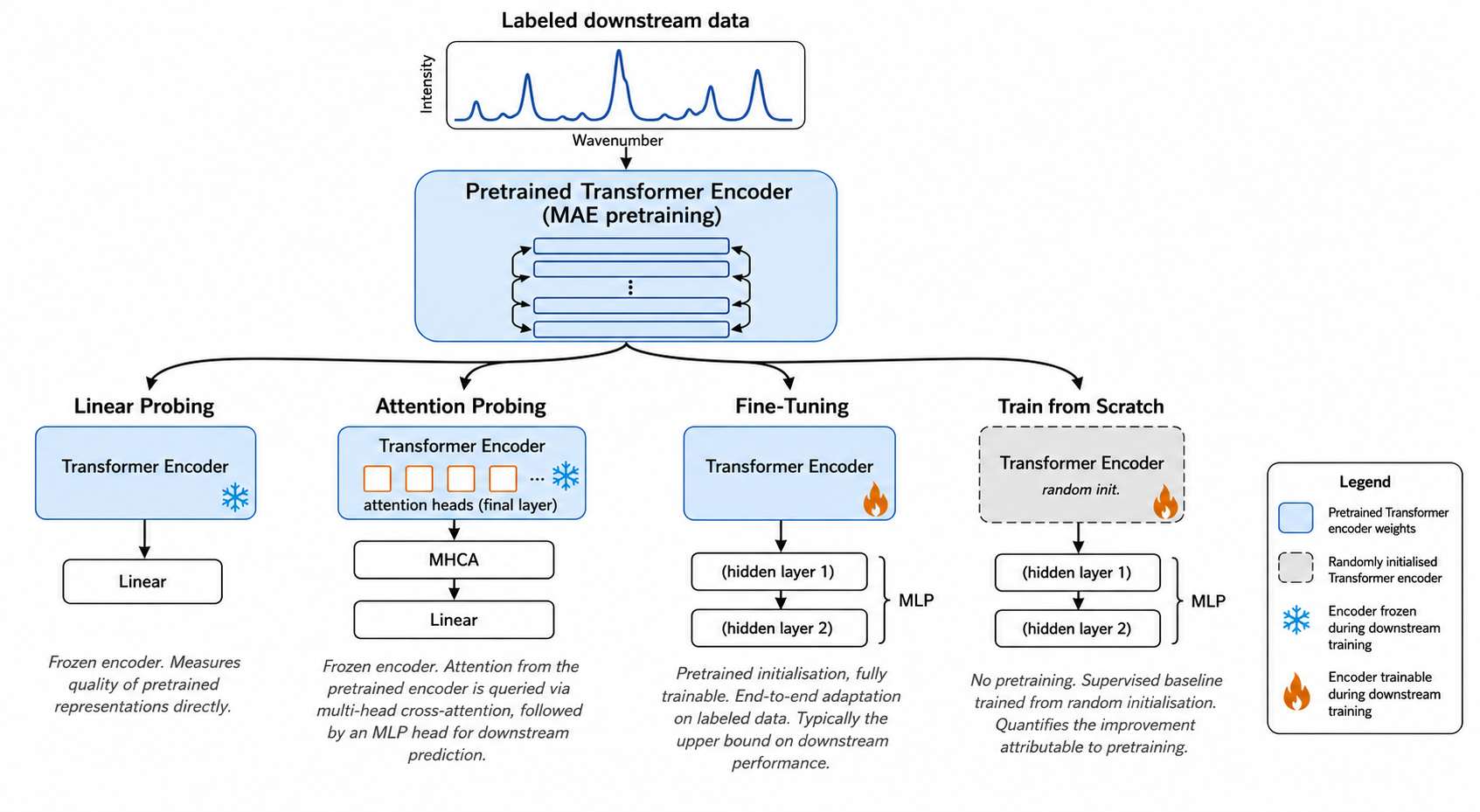}
    \caption{\textbf{Evaluation pipeline.} The four protocols used to
    assess the pretrained encoder, namely linear probing and attention probing
    (frozen encoder with a linear or attention-pooling head, respectively),
    fine-tuning (end-to-end adaptation), and training from scratch
    (random initialization).}
    \label{fig:eval}
\end{figure}

We evaluate representation quality through four complementary protocols: linear probing, attention probing, fine-tuning, and training from scratch. Together, these protocols assess different aspects of the learned representations, including feature quality without task adaptation, the effect of pretraining on optimization efficiency, and the interpretability of internal model behavior. Figure~\ref{fig:eval} illustrates the evaluation pipeline.

Linear probing is used to evaluate the quality of the learned representations when the encoder is frozen and only a linear classifier is trained on top of the [CLS] embedding using labeled data. This setting provides a strict measure of representation quality since no gradient updates are propagated into the encoder, and performance directly reflects the structure of the pretrained feature space.

Attention probing evaluates representation quality using a lightweight attention-pooling head trained on top of the frozen encoder's token embeddings, in place of the fixed [CLS] embedding used for linear probing. The probe learns a set of attention weights over the token sequence to produce a pooled representation, which is then passed to a classifier. Because the encoder itself remains frozen, this setting isolates the benefit of adaptive, task-specific pooling over the pretrained token representations from the benefit of full parameter adaptation, providing an intermediate evaluation point between linear probing and full fine-tuning.

In addition to probing-based evaluations, we consider full fine-tuning of the pretrained encoder on labeled data. In this setting, the encoder is initialized from pretraining and trained end-to-end, measuring the benefit of pretraining as an initialization strategy. This typically provides an upper bound on downstream performance for a fixed amount of labeled data.

Finally, we include a training-from-scratch baseline in which an architecture identical to the pretrained model is trained solely on labeled data with random initialization. This baseline quantifies the contribution of pretraining by comparing against models trained without any self-supervised signal.

\section{Results}
\label{sec:results}

\paragraph{Experimental setup.} Training uses AdamW with a learning rate of $1.5\times10^{-4}$, weight decay of $10^{-6}$, gradient clipping of $0.5$, and a cosine learning-rate schedule. Models trained from scratch are optimized for 50 epochs,  whereas linear probing, attention probing, and fine-tuning
are performed for 10 epochs. Bacterial classification is an exception:
all protocols use 30 epochs to match the training budget of the baseline
and ensure a fair comparison. The default masking ratio is 0.25, sinusoidal positional encodings are used unless otherwise stated, and the Voigt decoder predicts up to $K_{\max}=10$ candidate peaks. Spectra are normalized using min--max scaling, and dataset-specific sampling parameters are applied where appropriate. All experiments are repeated across three random seeds. 

Tables~\ref{tab:classification}, \ref{tab:Identification}, \ref{tab:concentration}, and \ref{tab:denoising} summarises performance across all downstream tasks and evaluation protocols for MAE-B; results for MAE-S and MAE-L and the effect of encoder scale are reported in Appendix~\ref{app:ablations}.
In Table~\ref{tab:classification}, the results demonstrate that the effectiveness of transfer learning is highly dependent on the characteristics of the downstream dataset.
For the Adenine dataset, which is essentially a binary classification task the benefit 
of fine-tuning is most prevalent. For the Bacteria dataset (30 classes), fine-tuning 
achieved a substantial improvement over both probing approaches and training from 
scratch, attaining an Accuracy score of approximately 66\% compared to 39\% for the randomly 
initialized model, note that the Bacteria dataset experiments have been conducted based on the held out dataset protocol as described in \cite{ho2019rapid}. Since the transfer learning experiments and the baseline study were 
both trained for 30 epochs, this comparison suggests that the pretrained representation 
provides a meaningful advantage under equivalent optimization conditions. The 
progressive improvement from linear probing to attention probing and full fine-tuning 
further indicates that while useful task-relevant information is present within the 
pretrained encoder, additional adaptation of the learned features is required to 
maximize performance on this task.

In contrast, all approaches produced comparable results on the binary classification 
Liver Cancer dataset, with F1 scores ranging from 88\% to 90\% and closely matching the 
baseline reported in the github repository from \cite{huang2023rapid}. The strong 
performance of the frozen encoder suggests that the pretrained representation already 
captures features relevant to the downstream classification task, while the limited 
gains from fine-tuning indicate that further parameter adaptation provides little 
additional benefit.

\begin{table}[t]
\centering
\caption{\textbf{Classification results (MAE-B).} Accuracy, AUC, AUPRC,
and F1-score (mean $\pm$ std over three seeds) on the Adenine, Bacteria,
Liver Cancer, and RRUFF-S (192- and 1332-class) datasets, for linear
probing, attention probing, fine-tuning, and training from scratch,
alongside published baselines where available.}
\vspace{4pt}
\label{tab:classification}
\makebox[\textwidth][c]{
\begin{tabular}{llcccc}
\toprule
\textbf{Dataset}  & \textbf{Evaluation} & \textbf{Accuracy} &
\textbf{AUC} & \textbf{AUPRC} & \textbf{F1-Score} \\

                 & \textbf{Protocols} &                  &
                 &                  &                    \\

\midrule
\multirow{4}{*}{Adenine} & Linear Probe & 0.6287 $\pm$ 0.0073 & 0.6613 $\pm$ 0.0188 & 0.6632 $\pm$ 0.0191 & 0.6060 $\pm$ 0.0170 \\
& Attention Probe & 0.7413 $\pm$ 0.0044 & 0.8100 $\pm$ 0.0086 & 0.7988 $\pm$ 0.0068 & 0.7009 $\pm$ 0.0063 \\

&  Fine Tune& 0.7787 $\pm$ 0.0017 & 0.8336 $\pm$ 0.0108 & 0.8238 $\pm$ 0.0067 & 0.7580 $\pm$ 0.0027 \\
& From Scratch & 0.7298 $\pm$ 0.0456 & 0.8170 $\pm$ 0.0425 & 0.8008 $\pm$ 0.0411 & 0.7351 $\pm$ 0.0453 \\

\hline

\multirow{4}{*}{Bacteria} & Linear Probe & 0.2639 $\pm$ 0.0265 & 0.8191 $\pm$ 0.0070 & 0.2537 $\pm$ 0.0170 & 0.2065 $\pm$ 0.0227 \\
 & Attention Probe & 0.4279 $\pm$ 0.0497 & 0.9193 $\pm$ 0.0109 & 0.4506 $\pm$ 0.0490 & 0.3987 $\pm$ 0.0548 \\
 & Fine Tune& 0.6611 $\pm$ 0.0133 & 0.9661 $\pm$ 0.0007 & 0.7061 $\pm$ 0.0083 & 0.6520 $\pm$ 0.0140 \\
 & From Scratch & $0.3938 \pm 0.0379$ & $0.9000\pm 0.0115$ & $0.4117 \pm 0.0333$ & $0.3319 \pm 0.0371$ \\
 &  All isolates baseline  & $0.8220 \pm 0.0030$ & - & - & - \\
\hline

       & Linear Probe & 0.8873 $\pm$ 0.0057 & 0.9492 $\pm$ 0.0053 & 0.9446 $\pm$ 0.0070 & 0.8873 $\pm$ 0.0057 \\
 Liver &  Attention Probe & 0.9053 $\pm$ 0.0112 & 0.9652 $\pm$ 0.0003 & 0.9650 $\pm$ 0.0007 & 0.9053 $\pm$ 0.0113 \\
 Cancer & Fine Tune& 0.8843 $\pm$ 0.0462 & 0.9609 $\pm$ 0.0006 & 0.9602 $\pm$ 0.0018 & 0.8836 $\pm$ 0.0474 \\
  & From Scratch & 0.8950 $\pm$ 0.0231 & 0.9590 $\pm$ 0.0016 & 0.9582 $\pm$ 0.0014 & 0.8948 $\pm$ 0.0232 \\
    &  Baseline & $\approx 0.9000$ & - & - & - \\
\hline

 & Linear Probe & 0.2558 $\pm$ 0.0132 & 0.9539 $\pm$ 0.0079 & 0.3903 $\pm$ 0.0484 & 0.1603 $\pm$ 0.0117 \\

 RRUFF-S &  Attention Probe & 0.4363 $\pm$ 0.0394 & 0.9786 $\pm$ 0.0044 & 0.6189 $\pm$ 0.0308 & 0.3313 $\pm$ 0.0447 \\
  (1/2) & Fine Tune& 0.9519 $\pm$ 0.0046 & 0.9997 $\pm$ 0.0001 & 0.9864 $\pm$ 0.0018 & 0.9479 $\pm$ 0.0057 \\
 & From Scratch & 0.9773 $\pm$ 0.0022 & 0.9979 $\pm$ 0.0035 & 0.9926 $\pm$ 0.0028 & 0.9766 $\pm$ 0.0021 \\
& Baseline & 0.9772 & 0.9979 & 0.9909 & 0.9763 \\
\hline

 & Linear Probe & 0.2055 $\pm$ 0.0011 & 0.9371 $\pm$ 0.0033 & 0.2872 $\pm$ 0.0071 & 0.1052 $\pm$ 0.0016 \\
  RRUFF-S & Attention Probe & 0.3136 $\pm$ 0.0219 & 0.9574 $\pm$ 0.0054 & 0.4283 $\pm$ 0.0288 & 0.1996 $\pm$ 0.0164 \\
  (2/2) & Fine Tune& 0.8454 $\pm$ 0.0083 & 0.9942 $\pm$ 0.0016 & 0.9282 $\pm$ 0.0085 & 0.8219 $\pm$ 0.0085 \\
 & From Scratch & 0.8439 $\pm$ 0.0166 & 0.6581 $\pm$ 0.0033 & 0.9235 $\pm$ 0.0137 & 0.8309 $\pm$ 0.0192 \\
 & Baseline & 0.8791 & 0.9691 & 0.9073 & 0.8511 \\
\bottomrule

\end{tabular}
}
\end{table}

The RRUFF-S datasets exhibit a different behavior. For the 192-class subset, training 
from scratch achieved the highest performance and closely matched the reported baseline, 
while fine-tuning resulted in a small performance degradation. Similarly, for the 1332-
class subset, fine-tuning and training from scratch produced nearly identical F1 scores, 
although both remained below the baseline. As the RRUFF-S experiments employed an 
approximately 69-70\% train and exactly 30\% train and test split respectively,  
consistent with the baseline study, these results suggest that the larger mineral 
spectroscopy datasets contain sufficient task-specific information to learn effective 
representations without relying heavily on pretraining. An interesting observation is 
the substantial difference in AUC between the fine-tuned and randomly initialized models 
on the 1332-class subset despite their comparable F1, precision, and accuracy scores. 
This behavior indicates that although both models achieve similar classification 
performance at the selected decision threshold, the fine-tuned model produces 
probability estimates with considerably better class separability across threshold 
values. Consequently, the pretrained representation appears to improve the ranking and 
calibration characteristics of the classifier even when threshold-dependent performance 
metrics remain largely unchanged.

Overall, the findings indicate that transfer learning provides the greatest benefit in 
data-constrained settings, whereas its advantage diminishes as the size and diversity of 
the downstream dataset increase. Furthermore, the results suggest that the benefits of 
pretraining may extend beyond improvements in classification accuracy alone, 
contributing to enhanced probability discrimination and confidence estimation in complex 
multiclass settings.

\begin{table}[t]
\centering
\caption{\textbf{Component identification results (MAE-B).} Accuracy,
AUC, AUPRC, and F1-score (mean $\pm$ std over three seeds) on the
DeepRaman (multi-component), DeepRaman (Liquid \& Powder), R6G, and Raman Sugars
datasets, for linear probing, attention probing, fine-tuning, and
training from scratch.}
\vspace{4pt}
\label{tab:Identification}
\makebox[\textwidth][c]{
\begin{tabular}{llcccc}
\toprule
\textbf{Dataset} & \textbf{Evaluation} & \textbf{Accuracy} &
\textbf{AUC} & \textbf{AUPRC} & \textbf{F1-Score} \\

             & \textbf{Protocol} &  &  &  &  \\
\midrule

                     & Linear Probe & $0.5903 \pm 0.0087$ & $0.6198 \pm 0.0020$ & $0.4359 \pm 0.0046$ & $0.5660 \pm 0.0110$ \\
   DeepRaman         & Attention Probe & $0.5975 \pm 0.0039$ & $0.5998 \pm 0.0202$ & $0.4171 \pm 0.0251$ & $0.5696 \pm 0.0048$ \\
  (Multi-            & Fine Tune & $0.5961 \pm 0.0014$ & $0.6075 \pm 0.0032$ & $0.4190 \pm 0.0107$ & $0.5729 \pm 0.0009$ \\
       component)    & From Scratch & $0.5953 \pm 0.0000$ & $0.6049 \pm 0.0336$ & $0.4271 \pm 0.0491$ & $0.5724 \pm 0.0000$ \\

\hline

          & Linear Probe & $0.8333 \pm 0.0000$ & $0.9200 \pm 0.0564$ & $0.8924 \pm 0.0867$ & $0.8500 \pm 0.0000$ \\
DeepRaman   & Attention Probe & $0.7222 \pm 0.1924$ & $0.7358 \pm 0.2207$ & $0.6860 \pm 0.1870$ & $0.7125 \pm 0.2382$ \\
(Liquid \&    & Fine Tune & $0.8333 \pm 0.0000$ & $0.8812 \pm 0.0387$ & $0.8653 \pm 0.0539$ & $0.8500 \pm 0.0000$ \\
Powder)            & From Scratch & $0.8333 \pm 0.0000$ & $0.7935 \pm 0.0143$ & $0.7373 \pm 0.0423$ & $0.8500 \pm 0.0000$ \\

\hline

\multirow{4}{*}{R6G}  & Linear Probe & $0.7634 \pm 0.0368$ & $0.9816 \pm 0.0012$ & $0.9823 \pm 0.0010$ & $0.7402 \pm 0.0451$ \\
                   & Attention Probe & $0,9979 \pm 0,0019$ & $0,9999 \pm 0,0001$ & $0,9999 \pm 0,0001$ & $0,9979 \pm 0,0019$ \\
                   & Fine Tune & $1.0000 \pm 0.0000$ & $1.0000 \pm 0.0000$ & $1.0000 \pm 0.0000$ & $1.0000 \pm 0.0000$ \\
                   & From Scratch & $0.9605 \pm 0.0350$ & $0.9993 \pm 0.0007$ & $0.9992 \pm 0.0007$ & $0.9599 \pm 0.0357$ \\

\hline

           & Linear Probe & $0.7449 \pm 0.0089$ & $0.6794 \pm 0.0051$ & $0.6495 \pm 0.0050$ & $0.5822 \pm 0.0186$ \\
    Raman  & Attention Probe & $0.8555 \pm 0.0437$ & $0.9187 \pm 0.0409$ & $0.8906 \pm 0.0499$ & $0.8171 \pm 0.0539$ \\
    Sugars & Fine Tune & $0.9523 \pm 0.0220$ & $0.9899 \pm 0.0060$ & $0.9855 \pm 0.0082$ & $0.9414 \pm 0.0251$ \\
           & From Scratch & $0.9519 \pm 0.0251$ & $0.9873 \pm 0.0087$ & $0.9822 \pm 0.0128$ & $0.9398 \pm 0.0312$ \\

\bottomrule
\end{tabular}
}
\end{table}

Table~\ref{tab:Identification} reports the results for the component identification task across four datasets of varying complexity. Overall, the results show that task difficulty is strongly influenced by the number of components present in each dataset, with performance generally improving as the number of components decreases.

The most challenging benchmark is the DeepRaman multi-component dataset, which contains 13 possible components. Across all evaluation protocols, the achieved performance remains relatively modest, with accuracies ranging from $0.5903$ to $0.5975$ and F1-scores ranging from $0.5660$ to $0.5729$. The similarity of the results obtained by linear probing, attention probing, fine-tuning, and training from scratch suggests that the complexity of the dataset limits the achievable performance regardless of the learning strategy. Fine-tuning achieves the highest F1-score of $0.5729$, while attention probing achieves the highest accuracy of $0.5975$. However, the differences between protocols remain small, indicating that the learned representations already capture most of the discriminative information available for this task.

For the Liquid \& Powder dataset, which involves identification among six components, performance improves substantially. Linear probing, fine-tuning, and training from scratch all achieve the highest accuracy of $0.8333$ and an F1-score of $0.8500$. However, linear probing obtains the best AUC and AUPRC values, reaching $0.9200$ and $0.8924$, respectively. Fine-tuning also performs strongly, with an AUC of $0.8812$ and an AUPRC of $0.8653$, outperforming training from scratch on these metrics. The attention probe shows lower average performance, with an accuracy of $0.7222$ and an F1-score of $0.7125$, suggesting reduced stability on this relatively small dataset.

The binary R6G identification task achieves strong overall performance, although the results vary substantially across evaluation protocols. Linear probing reaches an accuracy of $0.7634 \pm 0.0368$ and an F1-score of $0.7402 \pm 0.0451$, while maintaining high ranking performance, with an AUC of $0.9816 \pm 0.0012$ and an AUPRC of $0.9823 \pm 0.0010$. Attention probing substantially improves classification performance, achieving an accuracy and F1-score of $0.9979 \pm 0.0019$, together with an AUC and AUPRC of $0.9999 \pm 0.0001$. Fine-tuning achieves perfect and fully consistent results across all metrics, with accuracy, AUC, AUPRC, and F1-score values of $1.0000 \pm 0.0000$. Training from scratch also performs strongly, reaching an accuracy of $0.9605 \pm 0.0350$ and an F1-score of $0.9599 \pm 0.0357$, with an AUC of $0.9993 \pm 0.0007$ and an AUPRC of $0.9992 \pm 0.0007$. Overall, these findings indicate that the R6G spectral signatures are highly discriminative when the representation is adapted through attention probing or fine-tuning, whereas the lower linear-probing accuracy suggests that they are not perfectly separable using a simple linear classifier on fixed features.

The Raman Sugars dataset presents an intermediate level of difficulty in the component identification task. In this case, fine-tuning achieves the best overall performance, reaching an accuracy of $0.9523$, an AUC of $0.9899$, an AUPRC of $0.9855$, and an F1-score of $0.9414$. Training from scratch performs very similarly, with an accuracy of $0.9519$, an AUC of $0.9873$, an AUPRC of $0.9822$, and an F1-score of $0.9398$. Attention probing also performs strongly, substantially outperforming linear probing across all metrics, with an accuracy of $0.8555$ compared to $0.7449$ for linear probing, and an F1-score of $0.8171$ compared to $0.5822$. The significant improvement observed after fine-tuning suggests that task-specific adaptation is beneficial when subtle spectral differences exist between classes.

Across all datasets, a clear relationship emerges between the number of candidate components and classification performance. Datasets with fewer components generally yield higher accuracy, AUC, AUPRC, and F1-score values. Furthermore, pretrained approaches—particularly fine-tuning—generally match or exceed the performance of models trained from scratch, highlighting the effectiveness of transfer learning for Raman-based component identification. The largest benefits of fine-tuning are observed on datasets of intermediate difficulty, such as Raman Sugars, whereas for very simple tasks, such as R6G, or highly challenging tasks, such as DeepRaman multi-component identification, the differences between evaluation protocols become less pronounced.

Table \ref{tab:concentration} reports the results for the concentration estimation task across two Raman-based datasets with different numbers of target levels. The results show clear and consistent differences in the effectiveness of transfer learning depending on both dataset structure and supervision regime.

For the GBL dataset with four concentration levels, performance improves substantially as model capacity for adaptation increases. Linear probing achieves relatively weak performance, with an accuracy of 0.4689 and an F1-score of 0.2652, indicating limited separability under a frozen representation. Attention probing improves performance significantly, reaching an accuracy of 0.64 and an F1-score of 0.5331, suggesting that incorporating attention-based adaptation partially resolves representational limitations. Fine-tuning yields the best overall results, with an accuracy of 0.8733, AUC of 0.9814, and F1-score of 0.8351, demonstrating that full parameter adaptation is highly effective for this task. Training from scratch also performs strongly, achieving an F1-score of 0.7441 and an accuracy of 0.7297, although it remains consistently below the fine-tuned model. These results indicate that while pretraining provides a substantial advantage when fully leveraged through fine-tuning, learning from randomly initialized weights remains a competitive alternative for this dataset.

\begin{table}[t]
\centering
\caption{\textbf{Concentration estimation results (MAE-B).} Accuracy,
AUC, AUPRC, and F1-score (mean $\pm$ std over three seeds) on the GBL and
Raman Sugars datasets, framed as discrete-bin classification, for linear
probing, attention probing, fine-tuning, and training from scratch.}
\vspace{4pt}
\label{tab:concentration}
\makebox[\textwidth][c]{

\begin{tabular}{llcccc}
\toprule
\textbf{Dataset} & \textbf{Evaluation} & \textbf{Accuracy} &
\textbf{AUC} & \textbf{AUPRC} & \textbf{F1-Score} \\

 & \textbf{Protocol} & & & &  \\

\midrule

\multirow{4}{*}{GBL} & Linear Probe & $0.4689 \pm 0.0336$ & $0.7647 \pm 0.0227$ & $0.4428 \pm 0.0421$ & $0.2652 \pm 0.0526$ \\
                  & Attention Probe & $0.64 \pm 0.0115$ & $0.8916 \pm 0.0096$ & $0.7207 \pm 0.0576$ & $0.5331 \pm 0.0795$ \\
                  & Fine Tune & $0.8733 \pm 0.024$ & $0.9814 \pm 0.0042$ & $0.9359 \pm 0.0129$ & $0.8351 \pm 0.014$ \\
                  & From Scratch & $0.7297 \pm 0.0965$ & $0.9622 \pm 0.0178$ & $0.8903 \pm 0.0475$ & $0.7441 \pm 0.0811$ \\

\hline

           & Linear Probe & $0.3667 \pm 0.0419$ & $0.6781 \pm 0.0367$ & $0.4196 \pm 0.0288$ & $0.2201 \pm 0.0564$ \\
    Raman  & Attention Probe & $0.6764 \pm 0.1331$ & $0.7946 \pm 0.0641$ & $0.6268 \pm 0.0954$ & $0.5669 \pm 0.1111$ \\
    Sugars & Fine Tune & $0.8990 \pm 0.0206$ & $0.9275 \pm 0.0187$ & $0.9718 \pm 0.0038$ & $0.8883 \pm 0.0566$ \\
           & From Scratch & $0.9460 \pm 0.0163$ & $0.9007 \pm 0.0557$ & $0.9906 \pm 0.0013$ & $0.9461 \pm 0.0169$ \\

\bottomrule
\end{tabular}
}
\end{table}

For the Raman Sugars dataset with five concentration levels, a different trend emerges. Linear probing again yields the lowest performance, with an accuracy of 0.3667 and an F1-score of 0.2201. Attention probing improves results but remains limited compared to full training strategies. Fine-tuning achieves strong performance with an accuracy of 0.8990 and an F1-score of 0.8883, while training from scratch further improves accuracy to 0.9460 and achieves the highest F1-score of 0.9461. Notably, despite fine-tuning achieving high AUPRC (0.9718), training from scratch surpasses it across most metrics, suggesting that task-specific optimization from random initialization is particularly effective for this dataset.

Overall, the results indicate that concentration estimation benefits significantly from deeper model adaptation, with fine-tuning consistently outperforming shallow probing approaches. However, unlike the GBL dataset where pretraining provides the best overall performance when fully utilized, the Raman Sugars dataset demonstrates that training from scratch can match or exceed fine-tuning performance. This suggests that the usefulness of pretraining is highly dependent on the underlying data distribution and that, in some cases, task-specific learning from random initialization may yield superior discriminative performance.

Table~\ref{tab:denoising} summarizes denoising performance on the
MDA-Spectral dataset. Fine-tuning achieves an MSE of
$1.7996\times10^{-5}$ and a PSNR of 47.45 dB, compared with
$1.8797\times10^{-5}$ and 47.29 dB for training from scratch.
Conversely, training from scratch achieves a higher SSIM
(0.9603 versus 0.9575). These results show similar denoising
performance under the two initialization strategies, with
fine-tuning yielding slightly better reconstruction error and
training from scratch yielding slightly better structural similarity.

The published results of \citet{horgan2021high} are included for
reference. Their MSE measures spectral reconstruction error, whereas
their PSNR and SSIM assess reconstructed peak-intensity images.
Our PSNR and SSIM are evaluated directly on one-dimensional spectra.

\begin{table}[t]
\centering
\caption{\textbf{Denoising results (MAE-B) on MDA-Spectral.} MSE, PSNR,
and SSIM (mean $\pm$ std over three seeds) for fine-tuning and training
from scratch, compared against the published baseline.}
\vspace{4pt}
\label{tab:denoising}
\makebox[\textwidth][c]{
\begin{tabular}{llccc}
\toprule
\textbf{Dataset} & \textbf{Evaluation} & \textbf{MSE} & 
\textbf{PSNR} & \textbf{SSIM}   \\
 & \textbf{Protocol} & & & \\
\midrule	
\multirow{3}{*}{MDA-Spectral} & Fine Tune & $1.7996\times 10^{-5} \pm 0.1117\times 10^{-5}$ & $47.4535 \pm 0.2653$ & $0.9575 \pm 0.0015$ \\
                            & From Scratch & $1.8797\times 10^{-5} \pm 0.2864\times 10^{-5}$ & $47.2909 \pm 0.6374$ &  $0.9603 \pm 0.0001$ \\
                               & Baseline &  $2.85 \times 10^{-3}$ & 46.21 & 0.9532  \\
\bottomrule
\end{tabular}
}
\end{table}

The results indicate that pretraining does not provide the same level of advantage for the denoising task as observed for some of the classification datasets. Instead, both initialization strategies achieve comparable performance and significantly outperform the baseline method. This suggests that the MDA-Spectral denoising task may be sufficiently constrained for the network to learn robust reconstruction mappings directly from the available training data, while pretraining offers only modest improvements in reconstruction error and signal fidelity. Nevertheless, the lower MSE and higher PSNR obtained through fine-tuning indicate that pretraining can still provide a small but measurable benefit in optimizing reconstruction accuracy, even when overall denoising performance remains largely comparable between the two training strategies.

\section{Conclusion}
\label{sec:conclusion}

This article studied self-supervised pretraining for Raman spectroscopy.
We pretrained a masked autoencoder on unlabeled spectra from ten
biological, chemical, forensic, and materials datasets, using a pretext
task that combines per-pixel reconstruction, sparse Voigt-profile
reconstruction, and denoising. The Voigt decoder describes masked peak
regions with a few profiles whose centers, widths, and amplitudes are
interpretable as approximate peak parameters, while the per-pixel and
denoising objectives preserve dense signal fidelity.

We then evaluated the encoder across tasks, protocols, and scales, and
the outcome depends on the task. Fine-tuning the pretrained encoder improves on training the
same architecture from scratch where labeled data are scarce relative to
task difficulty, most clearly on 30-class bacterial identification, and
brings little or no benefit where downstream datasets are large or tasks
are near saturation. Frozen features are poorly served by a linear head
but considerably better by attention pooling. The value of pretraining
grows with encoder scale. Frozen features improve on most tasks, and on
several the margin over training from scratch widens, because larger
encoders trained from scratch gain less from added capacity and often
degrade. For
practitioners, this suggests starting from the pretrained encoder and
fine-tuning it when labels are limited, rather than relying on a linear
probe of frozen features.

Future work will follow the scaling trend observed here, pretraining
larger encoders on larger and more diverse corpora, and will evaluate
transfer to instruments and sample types held out from pretraining.

\section*{Acknowledgement}
This work has received funding from the European Union’s Horizon Innovation Actions under Grant Agreement no.101168416 (ARMADILLO).

\bibliographystyle{plainnat}
\bibliography{bibliography}

\appendix

\section{Voigt Profile Mathematics}
\label{app:voigt}

A Voigt profile is the convolution of a Lorentzian $\mathcal{L}$ and a Gaussian $\mathcal{G}$: \begin{equation} 
V(\nu; \nu_0, \Gamma_L, \Gamma_G) = \int_{-\infty}^{\infty} \mathcal{G}(\nu'; \Gamma_G)\, \mathcal{L}(\nu - \nu'; \nu_0, \Gamma_L)\, d\nu', 
\end{equation} 

where $\nu_0$ is the peak center (wavenumber), $\Gamma_L$ and $\Gamma_G$ are the Lorentzian and Gaussian half-widths at half-maximum (HWHM), respectively. For computational efficiency we use the Thompson \emph{et al.}\ pseudo-Voigt approximation \cite{thompson1987pvoigt}, which expresses the profile as a linear combination of a Lorentzian and a Gaussian with a mixing parameter $\eta(\Gamma_L, \Gamma_G) \in [0, 1]$: 

\begin{equation} V(\nu; A, \nu_0, \Gamma_L, \Gamma_G) \approx A \bigl[\eta\,\mathcal{L}(\nu; \nu_0, \Gamma) + (1 - \eta)\,\mathcal{G}(\nu; \nu_0, \Gamma)\bigr], 
\end{equation} 

where $A > 0$ is the amplitude, $\Gamma$ is the total HWHM, and $\eta$ is determined 
analytically from $\Gamma_L$ and $\Gamma_G$. This approximation is differentiable with 
respect to all parameters $(A, \nu_0, \Gamma_L, \Gamma_G)$, enabling end-to-end 
training. For each masked patch $i$, the 
Voigt head predicts a set of at most $K_{\max}$ candidate profiles. Each candidate $k 
\in \{1, \ldots, K_{\max}\}$ is associated with parameters $\boldsymbol{\theta}_k^{(i)} 
= (A_k, \nu_k, \Gamma_L^k, \Gamma_G^k)$ and a soft presence score $\alpha_k^{(i)} \in 
[0, 1]$ produced by a sigmoid output. The reconstructed spectrum over the masked patch 
region $\Omega_i$ is: 

\begin{equation} \hat{s}_{\text{Voigt}}^{(i)}(\nu) = \sum_{k=1}
^{K_{\max}} \alpha_k^{(i)}\, V\!\bigl(\nu;\, \boldsymbol{\theta}_k^{(i)}\bigr), \quad 
\nu \in \Omega_i. 
\end{equation} 

Physical constraints are enforced via parameter 
activations: $A_k$ and $\Gamma_L^k, \Gamma_G^k$ are passed through softplus to ensure 
positivity; $\nu_k$ is bounded to the patch wavenumber range by a sigmoid rescaling. 

The Voigt reconstruction loss combines spectral fidelity with a 
sparsity penalty on the number of active profiles: 

\begin{equation} \mathcal{L}
_{\text{Voigt}} = \frac{1}{|M|} \sum_{i \in M} \Bigl[ \lambda \underbrace{\frac{1}{N} 
\sum_{j=1}^{N} w_j^{(i)} \left( \hat{s}_{\text{Voigt},j}^{(i)} - x_{i,j} \right)^2}
_{\text{peak-weighted reconstruction}} + (1 - \lambda)\underbrace{\bigl\| \hat{s}
_{\text{Voigt}}^{(i)} - \mathbf{x}_i \bigr\|_2^2}_{\text{masked region reconstruction}} +\; 
\gamma \underbrace{\frac{1}{K_{\max}} \sum_{k=1}^{K_{\max}} \left| \alpha_k^{(i)} 
\right|}_{\text{sparsity}} \Bigr], \label{eq:voigt_loss} 
\end{equation} 

where 
\begin{equation} 
w_j^{(i)} = N \cdot \mathrm{softmax}\!\left(\frac{x_{i,j}}{T}\right). 
\end{equation} 

The sparsity term $\gamma \frac{1}{K_{\max}} \sum_k |\alpha_k^{(i)}|$ 
implements an $\ell_1$-type penalty on the presence scores, encouraging the model to 
explain each masked region with as few Voigt profiles as physically necessary. At 
inference time, profiles with $\alpha_k < \tau$ (threshold $\tau = 0.5$) are discarded, 
yielding an interpretable peak decomposition. Profiles with $
\alpha_k<\tau$ are discarded, yielding an interpretable decomposition of spectral peaks.

\section{Model Size Ablation}
\label{app:ablations}

The results reported in Section~\ref{sec:results} use the base model
configuration, MAE-B (Table~\ref{tab:model-scales}). To assess how
representation quality scales with model capacity, we repeat the full
evaluation protocol --- linear probing, attention probing, fine-tuning, and
training from scratch --- using the smaller MAE-S and larger MAE-L
configurations across all downstream tasks. Tables~\ref{tab:appendix_classification},
\ref{tab:appendix_Identification}, \ref{tab:appendix_concentration}, and
\ref{tab:appendix_denoising} report the corresponding results for the
classification, component identification, concentration estimation, and
denoising tasks, respectively, and should be read alongside the MAE-B
results in Section~\ref{sec:results} to assess the effect of encoder scale.

Two trends emerge across scales. First, the frozen representation
improves with encoder size. From MAE-S to MAE-L, linear-probe F1 rises by
0.11 to 0.57 on seven of the eleven prediction tasks (Adenine, Bacteria,
both RRUFF-S subsets, R6G, GBL, and Raman Sugars concentration), with the
largest gains on the 192-class RRUFF-S subset (0.05 to 0.62) and on
Bacteria (0.16 to 0.45). On the remaining four tasks, the two DeepRaman
datasets, Raman Sugars identification, and Liver Cancer, linear-probe F1
changes by at most 0.03. Second, the margin of fine-tuning over training
from scratch widens with encoder size on five tasks, by 0.05 to 0.22 F1:
both RRUFF-S subsets, Raman Sugars identification and concentration, and
GBL. On four of these five, the model trained from scratch degrades from
MAE-S to MAE-L, consistent with larger models overfitting the limited
labeled data when trained without pretraining, whereas the fine-tuned
model does not. The largest change occurs on the 1,332-class RRUFF-S
subset, where fine-tuning trails training from scratch by 0.12 F1 with
MAE-S but leads by 0.09 with MAE-L, and exceeds the published baseline
on all four metrics. Denoising follows the same pattern: fine-tuning
yields a higher MSE than training from scratch with MAE-S but a lower
one with MAE-L. The margin narrows by more than 0.02 F1 only on R6G
(by 0.07), where training from scratch itself approaches perfect
performance with MAE-L.

\begin{table}
\centering
\caption{\textbf{Classification results for MAE-S and MAE-L.} Accuracy,
AUC, AUPRC, and F1-score (mean $\pm$ std over three seeds) for all four
evaluation protocols across model scales, complementing the MAE-B
results in Table~\ref{tab:classification}.}
\vspace{4pt}
\label{tab:appendix_classification}
\makebox[\textwidth][c]{
\begin{tabular}{llcccc}
\toprule
\textbf{Dataset} & \textbf{Evaluation} & \textbf{Accuracy} &
\textbf{AUC} & \textbf{AUPRC} & \textbf{F1-Score} \\

                 & \textbf{Protocols} &                  &
                 &                  &                    \\

\midrule
\hline
\multicolumn{6}{c}{\textbf{MAE-SMALL}} \\
\hline

\multirow{4}{*}{Adenine} & Linear Probe & $0.5861 \pm 0.0185$ & $0.5953 \pm 0.0304$ & $0.5707 \pm 0.0489$ & $0.4734 \pm 0.0662$ \\
& Attention Probe & $0.6094 \pm 0.0419$ & $0.7313 \pm 0.0172$ & $0.7341 \pm 0.0149$ & $0.6078 \pm 0.0396$ \\
& Fine Tune & $0.7433 \pm 0.0093$ & $0.8294 \pm 0.0106$ & $0.8164 \pm 0.0092$ & $0.7260 \pm 0.0187$ \\
& From Scratch & $0.7291 \pm 0.0059$ & $0.8126 \pm 0.0073$ & $0.7924 \pm 0.0089$ & $0.7368 \pm 0.0063$ \\
\hline

\multirow{4}{*}{Bacteria} & Linear Probe & $0.2121 \pm 0.0307$ & $0.7816 \pm 0.0255$ & $0.2066 \pm 0.0098$ & $0.1623 \pm 0.0213$ \\
 & Attention Probe & $0.3233 \pm 0.0081$ & $0.8679 \pm 0.0050$ & $0.3329 \pm 0.0198$ & $0.2873 \pm 0.0079$ \\
 & Fine Tune & $0.6490 \pm 0.0094$ & $0.9709 \pm 0.0022$ & $0.6986 \pm 0.0179$ & $0.6404 \pm 0.0099$ \\
 & From Scratch & $0.3861 \pm 0.0115$ & $0.9116 \pm 0.0078$ & $0.4334 \pm 0.0011$ & $0.3246 \pm 0.0111$ \\
 & All isolates baseline  & $0.8220 \pm 0.0030$ & - & - & - \\
\hline

        & Linear Probe & $0.8560 \pm 0.0356$ & $0.9528 \pm 0.0017$ & $0.9515 \pm 0.0021$ & $0.8550 \pm 0.0370$ \\
 Liver  & Attention Probe & $0.8370 \pm 0.0035$ & $0.9611 \pm 0.0024$ & $0.9606 \pm 0.0022$ & $0.8350 \pm 0.0035$ \\
 Cancer & Fine Tune & $0.9083 \pm 0.0280$ & $0.9625 \pm 0.0018$ & $0.9609 \pm 0.0002$ & $0.9082 \pm 0.0282$ \\
        & From Scratch & $0.8970 \pm 0.0061$ & $0.9637 \pm 0.0041$ & $0.9633 \pm 0.0039$ & $0.8969 \pm 0.0061$ \\
        &  Baseline & $\approx 0.9000$ & - & - & - \\
\hline

         & Linear Probe & $0.1235 \pm 0.0282$ & $0.8797 \pm 0.0061$ & $0.1796 \pm 0.0213$ & $0.0509 \pm 0.0108$ \\
 RRUFF-S & Attention Probe & $0.1750 \pm 0.0197$ & $0.9219 \pm 0.0137$ & $0.2793 \pm 0.0370$ & $0.0768 \pm 0.0101$ \\
  (1/2)  & Fine Tune & $0.8889 \pm 0.0047$ & $0.9988 \pm 0.0005$ & $0.9591 \pm 0.0009$ & $0.8723 \pm 0.0029$ \\
         & From Scratch & $0.9793 \pm 0.0032$ & $0.9982 \pm 0.0030$ & $0.9928 \pm 0.0025$ & $0.9789 \pm 0.0032$ \\
         & Baseline & 0.9772 & 0.9979 & 0.9909 & 0.9763 \\
\hline

          & Linear Probe & $0.0841 \pm 0.0089$ & $0.8316 \pm 0.0085$ & $0.0856 \pm 0.0017$ & $0.0274 \pm 0.0055$ \\
  RRUFF-S & Attention Probe & $0.0990 \pm 0.0117$ & $0.8284 \pm 0.0138$ & $0.1107 \pm 0.0112$ & $0.0286 \pm 0.0036$ \\
  (2/2)   & Fine Tune & $0.7428 \pm 0.0014$ & $0.9949 \pm 0.0018$ & $0.8913 \pm 0.0045$ & $0.6939 \pm 0.0034$ \\
          & From Scratch & $0.8312 \pm 0.0172$ & $0.6576 \pm 0.0030$ & $0.9227 \pm 0.0068$ & $0.8163 \pm 0.0180$ \\
          & Baseline & 0.8791 & 0.9691 & 0.9073 & 0.8511 \\

\hline

\hline
\multicolumn{6}{c}{\textbf{MAE-LARGE}} \\
\hline

\multirow{4}{*}{Adenine} & Linear Probe & $0.6670 \pm 0.0117$ & $0.7249 \pm 0.0195$ & $0.7166 \pm 0.0106$ & $0.6449 \pm 0.0134$ \\
& Attention Probe & $0.7608 \pm 0.0111$ & $0.8031 \pm 0.0214$ & $0.7945 \pm 0.0200$ & $0.7332 \pm 0.0214$ \\
& Fine Tune & $0.7517 \pm 0.0052$ & $0.8150 \pm 0.0068$ & $0.8018 \pm 0.0064$ & $0.7176 \pm 0.0066$ \\
& From Scratch & $0.7278 \pm 0.0066$ & $0.8045 \pm 0.0096$ & $0.7965 \pm 0.0102$ & $0.7354 \pm 0.0072$ \\
\hline

\multirow{4}{*}{Bacteria} & Linear Probe & $0.4691 \pm 0.0067$ & $0.9165 \pm 0.0009$ & $0.4872 \pm 0.0067$ & $0.4485 \pm 0.0071$ \\
 & Attention Probe & $0.5694 \pm 0.0280$ & $0.9533 \pm 0.0041$ & $0.6129 \pm 0.0226$ & $0.5566 \pm 0.0306$ \\
 & Fine Tune & $0.6898 \pm 0.0059$ & $0.9644 \pm 0.0033$ & $0.7391 \pm 0.0090$ & $0.6797 \pm 0.0056$ \\
 & From Scratch & $0.4012 \pm 0.0284$ & $0.8701 \pm 0.0202$ & $0.4012 \pm 0.0407$ & $0.3467 \pm 0.0335$ \\
 & All isolates baseline  & $0.8220 \pm 0.0030$ & - & - & - \\
\hline

        & Linear Probe & $0.8317 \pm 0.0188$ & $0.9615 \pm 0.0030$ & $0.9603 \pm 0.0052$ & $0.8291 \pm 0.0203$ \\
 Liver  & Attention Probe & $0.8587 \pm 0.0601$ & $0.9599 \pm 0.0036$ & $0.9592 \pm 0.0039$ & $0.8566 \pm 0.0635$ \\
 Cancer & Fine Tune & $0.8703 \pm 0.0442$ & $0.9526 \pm 0.0122$ & $0.9518 \pm 0.0127$ & $0.8697 \pm 0.0448$ \\
        & From Scratch & $0.8740 \pm 0.0357$ & $0.9569 \pm 0.0088$ & $0.9549 \pm 0.0103$ & $0.8735 \pm 0.0363$ \\
        &  Baseline & $\approx 0.9000$ & - & - & - \\

\hline

         & Linear Probe & $0.6832 \pm 0.0142$ & $0.9955 \pm 0.0010$ & $0.8486 \pm 0.0087$ & $0.6223 \pm 0.0182$ \\
 RRUFF-S & Attention Probe & $0.8176 \pm 0.0151$ & $0.9965 \pm 0.0009$ & $0.9071 \pm 0.0087$ & $0.7938 \pm 0.0174$ \\
  (1/2)  & Fine Tune & $0.9612 \pm 0.0095$ & $0.9999 \pm 0.0000$ & $0.9910 \pm 0.0011$ & $0.9594 \pm 0.0109$ \\
         & From Scratch & $0.9690 \pm 0.0053$ & $0.9981 \pm 0.0029$ & $0.9892 \pm 0.0016$ & $0.9696 \pm 0.0025$ \\
         & Baseline & 0.9772 & 0.9979 & 0.9909 & 0.9763 \\

\hline
          & Linear Probe & $0.4751 \pm 0.0020$ & $0.9862 \pm 0.0012$ & $0.7295 \pm 0.0047$ & $0.3598 \pm 0.0042$ \\
  RRUFF-S & Attention Probe & $0.5721 \pm 0.0316$ & $0.9871 \pm 0.0015$ & $0.7606 \pm 0.0248$ & $0.4915 \pm 0.0358$ \\
  (2/2)   & Fine Tune & $0.8939 \pm 0.0074$ & $0.9962 \pm 0.0009$ & $0.9507 \pm 0.0063$ & $0.8800 \pm 0.0109$ \\
          & From Scratch & $0.8054 \pm 0.0249$ & $0.6555 \pm 0.0047$ & $0.9092 \pm 0.0045$ & $0.7872 \pm 0.0275$ \\
          & Baseline & 0.8791 & 0.9691 & 0.9073 & 0.8511 \\

\hline

\end{tabular}
}
\end{table}

\begin{table}
\centering
\caption{\textbf{Component identification results for MAE-S and MAE-L.}
Accuracy, AUC, AUPRC, and F1-score (mean $\pm$ std over three seeds) for
all four evaluation protocols across model scales, complementing the
MAE-B results in Table~\ref{tab:Identification}.}
\vspace{4pt}
\label{tab:appendix_Identification}
\makebox[\textwidth][c]{
\begin{tabular}{llcccc}
\toprule
\textbf{Dataset} & \textbf{Evaluation} & \textbf{Accuracy} &
\textbf{AUC} & \textbf{AUPRC} & \textbf{F1-Score} \\

             & \textbf{Protocol} &  &  &  &  \\
\midrule
\hline
\multicolumn{6}{c}{\textbf{MAE-SMALL}} \\
\hline

                     & Linear Probe & $0.5945 \pm 0.0014$ & $0.6463 \pm 0.0221$ & $0.4886 \pm 0.0387$ & $0.5653 \pm 0.0123$ \\
   DeepRaman         & Attention Probe & $0.5886 \pm 0.0347$ & $0.4763 \pm 0.0176$ & $0.4588 \pm 0.1588$ & $0.2518 \pm 0.2885$ \\
  (Multi-            & Fine Tune & $0.5914 \pm 0.0068$ & $0.6056 \pm 0.0454$ & $0.4246 \pm 0.0459$ & $0.5615 \pm 0.0189$ \\
       component)    & From Scratch & $0.5953 \pm 0.0000$ & $0.5935 \pm 0.0234$ & $0.4004 \pm 0.0193$ & $0.5724 \pm 0.0000$ \\

\hline

         & Linear Probe & $0.8333 \pm 0.0000$ & $0.8121 \pm 0.0686$ & $0.7472 \pm 0.1428$ & $0.8500 \pm 0.0000$ \\
DeepRaman   & Attention Probe & $0.7314 \pm 0.1764$ & $0.6904 \pm 0.1657$ & $0.6207 \pm 0.1348$ & $0.7333 \pm 0.2021$ \\
(Liquid \&      & Fine Tune & $0.8333 \pm 0.0000$ & $0.8446 \pm 0.0386$ & $0.7949 \pm 0.0549$ & $0.8500 \pm 0.0000$ \\
Powder)           & From Scratch & $0.8333 \pm 0.0000$ & $0.7956 \pm 0.0697$ & $0.6957 \pm 0.1064$ & $0.8500 \pm 0.0000$ \\

\hline
\multirow{4}{*}{R6G}  & Linear Probe & $0,7131 \pm 0,0175$ & $0,9768 \pm 0,0032$ & $0,9778 \pm 0,0031$ & $0,6756 \pm 0,0234$ \\
                  & Attention Probe & $0,8885 \pm 0,0489$ & $0,9897 \pm 0,0087$ & $0,9899 \pm 0,0085$ & $0,8854 \pm 0,0518$ \\
                  & Fine Tune & $1,0000 \pm 0,0000$ & $1,0000 \pm 0,0000$ & $1,0000 \pm 0,0000$ & $1,0000 \pm 0,0000$ \\
                  & From Scratch & $0,9239 \pm 0,0489$ & $0,9758 \pm 0,0238$ & $0,9771 \pm 0,0224$ & $0,9238 \pm 0,0489$ \\

\hline

           & Linear Probe & $0,7463 \pm 0,0045$ & $0,6808 \pm 0,0123$ & $0,6460 \pm 0,0091$ & $0,5571 \pm 0,0363$ \\
    Raman  & Attention Probe & $0,7614 \pm 0,0205$ & $0,7430 \pm 0,0517$ & $0,7109 \pm 0,0488$ & $0,6280 \pm 0,0467$ \\
    Sugars & Fine Tune & $0,9391 \pm 0,0066$ & $0,9825 \pm 0,0024$ & $0,9742 \pm 0,0035$ & $0,9219 \pm 0,0078$ \\
           & From Scratch & $0,9195 \pm 0,0572$ & $0,9658 \pm 0,0336$ & $0,9546 \pm 0,0394$ & $0,8933 \pm 0,0785$ \\
\hline

\hline
\multicolumn{6}{c}{\textbf{MAE-LARGE}} \\
\hline
                     & Linear Probe & $0.6059 \pm 0.0128$ & $0.6422 \pm 0.0200$ & $0.4644 \pm 0.0545$ & $0.5721 \pm 0.0040$ \\
   DeepRaman         & Attention Probe & $0.5936 \pm 0.0029$ & $0.6041 \pm 0.0267$ & $0.4203 \pm 0.0316$ & $0.5557 \pm 0.0289$ \\
  (Multi-            & Fine Tune & $0.5958 \pm 0.0009$ & $0.6025 \pm 0.0156$ & $0.4099 \pm 0.0244$ & $0.5727 \pm 0.0006$ \\
       component)    & From Scratch & $0.5953 \pm 0.0000$ & $0.6100 \pm 0.0347$ & $0.4342 \pm 0.0425$ & $0.5724 \pm 0.0000$ \\

\hline

          & Linear Probe & $0.8333 \pm 0.0000$ & $0.8090 \pm 0.0533$ & $0.7349 \pm 0.1103$ & $0.8500 \pm 0.0000$ \\
DeepRaman   & Attention Probe & $0.7777 \pm 0.0962$ & $0.7647 \pm 0.0891$ & $0.6886 \pm 0.0424$ & $0.7571 \pm 0.1608$ \\
(Liquid \&     & Fine Tune & $0.8333 \pm 0.0000$ & $0.8590 \pm 0.0901$ & $0.8049 \pm 0.1396$ & $0.8500 \pm 0.0000$ \\
Powder)           & From Scratch & $0.8333 \pm 0.0000$ & $0.8937 \pm 0.0143$ & $0.8688 \pm 0.0212$ & $0.8500 \pm 0.0000$ \\

\hline
\multirow{4}{*}{R6G}  & Linear Probe & $0,8022 \pm 0,0262$ & $0,9578 \pm 0,0007$ & $0,9602 \pm 0,0005$ & $0,7887 \pm 0,0315$ \\
                  & Attention Probe & $1,0000 \pm 0,0000$ & $1,0000 \pm 0,0000$ & $1,0000 \pm 0,0000$ & $1,0000 \pm 0,0000$ \\
                  & Fine Tune & $1,0000 \pm 0,0000$ & $1,0000 \pm 0,0000$ & $1,0000 \pm 0,0000$ & $1,0000 \pm 0,0000$ \\
                  & From Scratch & $0,9982 \pm 0,0016$ & $0,9998 \pm 0,0002$ & $0,9998 \pm 0,0002$ & $0,9982 \pm 0,0016$ \\

\hline

           & Linear Probe & $0,7419 \pm 0,0068$ & $0,6941 \pm 0,0063$ & $0,6475 \pm 0,0050$ & $0,5443 \pm 0,0436$ \\
    Raman  & Attention Probe & $0,9227 \pm 0,0178$ & $0,9739 \pm 0,0099$ & $0,9610 \pm 0,0131$ & $0,9011 \pm 0,0187$ \\
    Sugars & Fine Tune & $0,9557 \pm 0,0215$ & $0,9895 \pm 0,0076$ & $0,9842 \pm 0,0104$ & $0,9438 \pm 0,0275$ \\
           & From Scratch & $0,8964 \pm 0,0708$ & $0,9491 \pm 0,0434$ & $0,9343 \pm 0,0561$ & $0,8648 \pm 0,0930$ \\
\bottomrule

\end{tabular}
}
\end{table}

\begin{table}
\centering
\caption{\textbf{Concentration estimation results for MAE-S and MAE-L.}
Accuracy, AUC, AUPRC, and F1-score (mean $\pm$ std over three seeds) for
all four evaluation protocols across model scales, complementing the
MAE-B results in Table~\ref{tab:concentration}.}
\vspace{4pt}
\label{tab:appendix_concentration}
\makebox[\textwidth][c]{

\begin{tabular}{llcccc}
\toprule
\textbf{Dataset} & \textbf{Evaluation} & \textbf{Accuracy} &
\textbf{AUC} & \textbf{AUPRC} & \textbf{F1-Score} \\

  & \textbf{Protocol} & & & &  \\

\midrule
\hline
\multicolumn{6}{c}{\textbf{MAE-SMALL}} \\
\hline

\multirow{4}{*}{GBL}  &  Linear Probe & $0.4511 \pm 0.0252$ & $0.6965 \pm 0.0019$ & $0.3794 \pm 0.0269$ & $0.1914 \pm 0.0344$ \\
                      &  Attention Probe & $0.5311 \pm 0.0391$ & $0.8084 \pm 0.0297$ & $0.5194 \pm 0.0551$ & $0.2998 \pm 0.0920$ \\
                      &  Fine Tune & $0.8022 \pm 0.0308$ & $0.9586 \pm 0.0102$ & $0.8496 \pm 0.0436$ & $0.7445 \pm 0.0649$ \\
                      &  From Scratch & $0.8220 \pm 0.0404$ & $0.9756 \pm 0.0040$ & $0.9128 \pm 0.0046$ & $0.8094 \pm 0.0393$ \\
\hline

           & Linear Probe & $0.2838 \pm 0.0148$ & $0.6897 \pm 0.0077$ & $0.4074 \pm 0.0050$ & $0.1198 \pm 0.0252$ \\
    Raman  & Attention Probe & $0.4478 \pm 0.0188$ & $0.6896 \pm 0.0134$ & $0.4291 \pm 0.0233$ & $0.3487 \pm 0.0258$ \\
    Sugars & Fine Tune & $0.8744 \pm 0.0094$ & $0.9264 \pm 0.0202$ & $0.8483 \pm 0.0680$ & $0.7629 \pm 0.0314$ \\
           & From Scratch & $0.9558 \pm 0.0156$ & $0.9614 \pm 0.0103$ & $0.9861 \pm 0.0037$ & $0.9527 \pm 0.0120$ \\
\hline

\hline
\multicolumn{6}{c}{\textbf{MAE-LARGE}} \\
\hline

\multirow{4}{*}{GBL}  &  Linear Probe & $0.5133 \pm 0.0437$ & $0.7604 \pm 0.0359$ & $0.5264 \pm 0.0886$ & $0.372 \pm 0.0992$ \\
                  & Attention Probe & $0.6422 \pm 0.0543$ & $0.87 \pm 0.0198$ & $0.7179 \pm 0.0455$ & $0.6113 \pm 0.0863$ \\
                  & Fine Tune & $0.7267 \pm 0.0416$ & $0.9413 \pm 0.0034$ & $0.8855 \pm 0.0109$ & $0.7466 \pm 0.046$ \\
                  & From Scratch & $0.7166 \pm 0.0588$ & $0.9052 \pm 0.0211$ & $0.8032 \pm 0.0577$ & $0.701 \pm 0.0739$ \\

\hline

           &  Linear Probe & $0.4880 \pm 0.0042$ & $0.7579 \pm 0.0204$ & $0.4931 \pm 0.0296$ & $0.4138 \pm 0.0171$ \\
    Raman  &  Attention Probe & $0.7896 \pm 0.0189$ & $0.8668 \pm 0.0363$ & $0.7971 \pm 0.0547$ & $0.7258 \pm 0.0696$ \\
    Sugars &  Fine Tune & $0.9162 \pm 0.0102$ & $0.9178 \pm 0.0438$ & $0.9789 \pm 0.0053$ & $0.9260 \pm 0.0064$ \\
           &  From Scratch & $0.9835 \pm 0.0045$ & $0.9559 \pm 0.0323$ & $0.9966 \pm 0.0011$ & $0.9821 \pm 0.0036$ \\

\bottomrule

\end{tabular}
}
\end{table}

\begin{table}
\centering
\caption{\textbf{Denoising results on MDA-Spectral for MAE-S and
MAE-L.} MSE, PSNR, and SSIM (mean $\pm$ std over three seeds) for
fine-tuning and training from scratch across model scales, compared
against the published baseline and complementing the MAE-B results in
Table~\ref{tab:denoising}.}
\vspace{4pt}
\label{tab:appendix_denoising}
\makebox[\textwidth][c]{
\begin{tabular}{llccc}
\toprule
\textbf{Dataset} & \textbf{Evaluation} & \textbf{MSE} & 
\textbf{PSNR} & \textbf{SSIM}   \\
 & \textbf{Protocol} & & & \\
\midrule
\hline

\multicolumn{5}{c}{\textbf{MAE-SMALL}} \\
\hline
\multirow{3}{*}{MDA-Spectral}  & Fine Tune & $1.7510\times 10^{-5} \pm 0.0705\times 10^{-5}$ & $47.5364 \pm 0.1795$ & $0.9597 \pm 0.0003$ \\	
                               & From Scratch & $1.6144\times 10^{-5} \pm 0.0887\times 10^{-5}$ & $47.7670 \pm 0.2270$ & $0.9608 \pm 0.0002$ \\
                               & Baseline & $2.85 \times 10^{-3}$ & 46.21 & 0.9532  \\
\hline
\hline
\multicolumn{5}{c}{\textbf{MAE-LARGE}} \\
	
\hline
\multirow{3}{*}{MDA-Spectral}  & Fine Tune & $1.7052\times 10^{-5} \pm 0.0520\times 10^{-5}$ & $47.6834 \pm 0.1317$& $0.9594 \pm 0.0019$ \\
                               & From Scratch & $2.0057\times 10^{-5}\pm 0.1418\times 10^{-5}$ &  $46.9844\pm 0.3076$ & $0.9603\pm 0.0003$ \\
                               & Baseline & $2.85 \times 10^{-3}$ & 46.21 & 0.9532  \\
\bottomrule
\end{tabular}
}
\end{table}

\section{Qualitative Analysis of Maximum Peak Reconstruction}

The Voigt reconstruction head is designed to provide a compact representation of localized Raman peak structure. Rather than predicting every intensity value independently, the decoder represents a masked spectral region through a small set of parameterized Voigt profiles. Each profile is defined by its amplitude, center position, Gaussian width, Lorentzian width, and presence score. This parameterization allows the model to summarize the dominant morphology of a peak using substantially fewer quantities than direct per-pixel reconstruction.

The objective is not to assume that every Raman band follows an ideal Voigt line shape. Instead, the Voigt function is used as a general representation of peak-like spectral structure. Its parameters capture the principal properties of a localized band, including its position, intensity, and width, while the sparsity constraint encourages the reconstruction to rely on a limited number of active components.

\begin{figure}
    \centering
    \includegraphics[width=0.9\linewidth, trim={0 50 0 32}, clip]{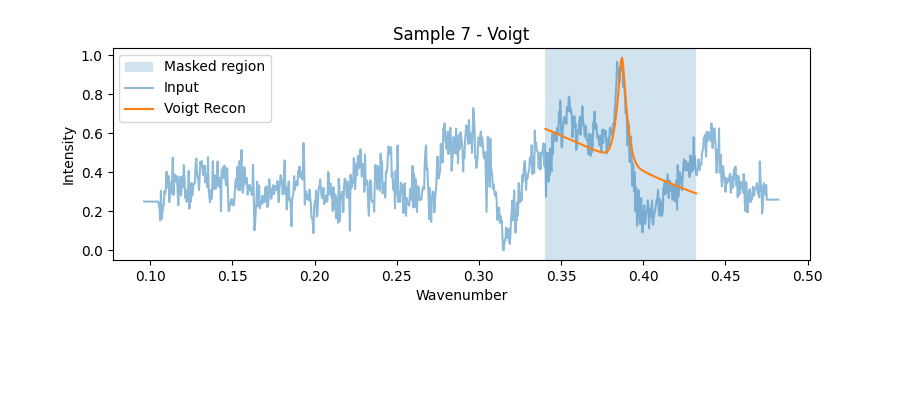}
    \caption{Representative Voigt-based reconstruction of a Raman peak. The input spectrum is shown together with the reconstruction produced by the sparse Voigt decoder. The reconstructed profile captures the location and overall morphology of the dominant peak using a compact parameterized representation.}
    \label{fig:voigt_reconstruction_example}
\end{figure}

Figure~\ref{fig:voigt_reconstruction_example} presents a representative reconstruction produced by the Voigt branch. The predicted profile captures the position and overall width of the dominant Raman peak while producing a smooth approximation of the surrounding signal. In comparison with an unconstrained per-pixel decoder, the Voigt representation suppresses small point-wise fluctuations and emphasizes the broader morphology of the peak. This behavior is consistent with the role of the Voigt branch as a structured compression mechanism rather than as a direct replacement for dense spectral reconstruction.

The qualitative example illustrates that the decoder can approximate a localized Raman feature using a compact set of peak parameters. The resulting reconstruction is smoother than the original spectrum and does not reproduce every small fluctuation. This is expected, as the branch is intended to preserve the dominant peak structure rather than model all local noise and baseline variation. The per-pixel and Voigt branches therefore play complementary roles: the per-pixel branch preserves detailed signal intensity, whereas the Voigt branch encourages the encoder to retain a low-dimensional description of the underlying peak morphology.

The per-pixel reconstruction branch predicts the intensity values of the masked spectral region directly. Unlike the Voigt branch, it is not constrained to represent the signal using a particular peak family and can therefore model both localized peaks and broader spectral variation. This branch provides the standard masked-autoencoding objective and encourages the encoder to retain sufficient contextual information to recover the missing signal.

\begin{figure}
    \centering
    \includegraphics[width=0.9\linewidth, trim={0 50 0 32}, clip]{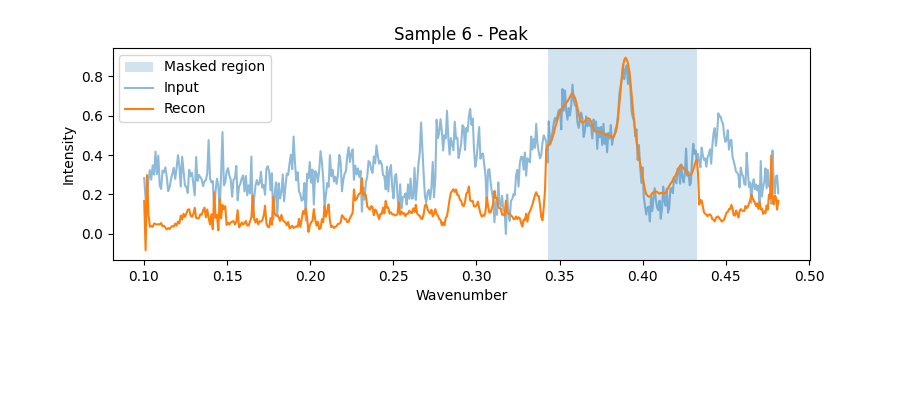}
    \caption{Representative per-pixel reconstruction. The decoder recovers the overall spectral shape and the dominant Raman features through direct prediction of the intensity values.}
    \label{fig:pixel_reconstruction_example}
\end{figure}

Figure~\ref{fig:pixel_reconstruction_example} shows an example of a spectrum reconstructed by the per-pixel decoder. The prediction follows the overall signal shape and recovers the dominant peak locations, while smaller fluctuations and fine-scale intensity variations are reproduced less precisely. The reconstruction demonstrates that the encoder retains information about both the global spectral trend and the principal local structures required to estimate the missing region.

The per-pixel and Voigt reconstructions should not be expected to behave identically. The per-pixel decoder has greater flexibility and can approximate arbitrary signal shapes, while the Voigt decoder is restricted to a compact combination of smooth peak profiles. The two branches therefore provide different training signals. Dense reconstruction encourages local fidelity, whereas Voigt reconstruction encourages the representation of peak position, amplitude, and width through a lower-dimensional parameterization.

\end{document}